%% file: main.tex
\documentclass{article}

\usepackage{microtype}
\usepackage{graphicx}
\usepackage{subfigure}
\usepackage{booktabs} %

\usepackage{hyperref}

\usepackage[accepted]{icml2025}

\usepackage{amsmath}
\usepackage{amssymb}
\usepackage{mathtools}
\usepackage{amsthm}

\usepackage[capitalize,noabbrev]{cleveref}

\theoremstyle{plain}

\theoremstyle{definition}

\theoremstyle{remark}

\usepackage[textsize=tiny]{todonotes}

\icmltitlerunning{SMAAT}
\usepackage{algorithm}
\usepackage{algorithmic}

\newcounter{footnotemarknum}
\newcommand{\fnm}{\addtocounter{footnotemarknum}{1}\footnotemark} 
\newcommand{\fnmNon}{\addtocounter{footnote}{-1}\footnotemark} 

\newcommand{\fnt}[1]{
    \addtocounter{footnote}{-\value{footnotemarknum}}
    \addtocounter{footnote}{1}
    \footnotetext{#1}
    \setcounter{footnotemarknum}{0}
}

\usepackage{microtype}
\usepackage{graphicx}
\usepackage{subfigure}
\usepackage{amsfonts}
\usepackage{amssymb}
\usepackage{booktabs} %
\usepackage{amsmath} 
\usepackage{multirow}
\usepackage{xcolor}
\usepackage{eqnarray}
\usepackage{mathtools}

\usepackage{xcolor}

\usepackage{tablefootnote}

\renewcommand*\cite[1]{\citep{#1}}

\usepackage{nicematrix}

\usepackage{bm}
\usepackage{multirow}

\newcommand{\figref}[1]{Fig.~\ref{#1}}

\newcommand{\cD}{\mathcal{D}}
\newcommand{\cO}{\mathcal{O}}

\newcommand{\cY}{\mathcal{\vect Y}}

\newcommand{\cX}{\mathcal{\vect X}}

\newcommand{\vect}[1]{\boldsymbol{#1}}

\usepackage{xspace}

\def\eg{\emph{e.g.,} } 
\def\ie{\emph{i.e.,} }

\usepackage{mathtools}
\usepackage{tcolorbox}

\usepackage{xspace}
\newcommand{\system}{\texttt{SMAAT}}

\begin{document}

\twocolumn[
\icmltitle{Explaining the role of Intrinsic Dimensionality in %
Adversarial Training }

\begin{icmlauthorlist}
\icmlauthor{Enes Altinisik}{QCRI}
\icmlauthor{Safa Messaoud}{QCRI}
\icmlauthor{Husrev Taha Sencar}{QCRI}
\icmlauthor{Hassan Sajjad}{Dalhouse}
\icmlauthor{Sanjay Chawla}{QCRI}
\end{icmlauthorlist}

\icmlaffiliation{QCRI}{Qatar Computing Research Institute, HBKU, Doha, Qatar}
\icmlaffiliation{Dalhouse}{Faculty of Computer Science,Dalhousie University, Halifax, Canada}

\icmlcorrespondingauthor{Enes Altinisik}{ealtinisik@hbku.edu.qa}

\icmlkeywords{Machine Learning, ICML}
\vskip 0.3in
]

\printAffiliationsAndNotice{}

\begin{abstract}

Adversarial Training (AT) impacts different architectures in distinct ways: vision models gain robustness but face reduced generalization, encoder-based models exhibit limited robustness improvements with minimal generalization loss, and recent work in latent-space adversarial training (LAT) demonstrates that decoder-based models achieve improved robustness by applying AT across multiple layers.
We provide the first explanation for these trends by leveraging the manifold conjecture: off-manifold adversarial examples (AEs) enhance robustness, while on-manifold AEs improve generalization.
We show that vision and decoder-based models exhibit low intrinsic dimensionality in earlier layers (favoring off-manifold AEs), whereas encoder-based models do so in later layers (favoring on-manifold AEs). 
Exploiting this property, we introduce \system{}, which improves the scalability of AT for encoder-based models by perturbing the layer with the lowest intrinsic dimensionality. This reduces the projected gradient descent (PGD) chain length required for AE generation, cutting GPU time by 25–33\% while significantly boosting robustness. We validate \system{} across multiple tasks, including text generation, sentiment classification, safety filtering, and retrieval augmented generation setups, demonstrating superior robustness with comparable generalization to standard training.

\end{abstract}

\input{introduction}

\input{related_work}

\input{approach}

\input{experiments}

\input{conclusion}

\section*{Acknowledgement}
This work is partial supported by the Qatar National Research Fund (QNRF) grant NPRP11C-1229-170007.

\input{impact}

\bibliography{custom}
\bibliographystyle{icml2025}

\appendix
\input{appendix}

\end{document}

%% file: introduction.tex
\section{Introduction}
\label{introduction}
\input{figures/llama_teaser}

Adversarial Training (AT) is the most effective approach for improving the robustness of deep neural networks against small input perturbations~\cite{ijcai2021p591,kurakin2018adversarial}. It is formulated as a min-max optimization problem, where the outer minimization optimizes model parameters, and the inner maximization seeks worst-case input perturbations. In deep networks, the inner maximization is typically solved approximately using multiple iterations of projected gradient descent (PGD, \citet{madry2017towards}). 
However, the trade-off between robustness and generalization remains poorly understood. For instance, AT reduces generalization in vision models, leading to an 8\% drop on CIFAR-10~\cite{shafahi2019adversarial}, whereas encoder-based language models often retain or even improve generalization, achieving a 1\% gain on AGNEWS~\cite{Zhu2020FreeLB}. 
Moreover, robustness gains are significantly higher in vision models (e.g., 40\% on CIFAR-10) compared to encoder-based models (e.g., 10\% on AGNEWS). 
Recent work on Latent Adversarial Training (LAT) suggests that applying AT across multiple layers yields better robustness and generalization than focusing on a single layer~\cite{sheshadri2024latent}. While prior research has investigated generalization loss in vision models~\cite{madry2017towards,wang2019improving,altinisik2023a3t,zhang2019theoretically,cheng2020cat}, no study has systematically explored the robustness-generalization dynamics across vision models (CNNs and Vision Transformers), decoder-based (dec-LLMs), and encoder-based (enc-LLMs) language models. 
Additionally, the high computational cost of AT limits its practical deployment. Recent efforts to reduce the number of PGD steps—by reusing or accumulating gradients during updates~\cite{shafahi2019adversarial,zhang2019you,Zhu2020FreeLB}—have improved efficiency but still require full network passes, making AT computationally expensive.

\input{figures/fig_intor_compare}
In this work, we investigate how differences in robustness and generalization trends across foundational models relate to the intrinsic dimensionality (ID) of the data manifold. 
The data manifold is a potentially non-linear subspace spanned by the dataset, and its dimensionality influences whether adversarial examples (AEs) lie on the manifold (on-manifold AEs) or fall outside it (off-manifold AEs) during training.
We show that the ID of the data manifold in the first layer is much higher in enc-LLMs compared to vision models and dec-LLMs  
which results in AEs being more on-manifold (ONM-AEs) in enc-LLMs and more off-manifold (OFM-AEs) in vision models and dec-LLMs. 
In accordance with the manifold conjecture~\cite{ethayarajh-2019-contextual, shamir2021dimpled, DBLP:conf/iclr/GilmerMFSRWG18}, we find that OFM-AEs lead to better robustness, while more ONM-AEs lead to better generalization (Fig. \ref{fig:teaser}, with details provided in Section \ref{seq:approach}). To the best of our knowledge, this is the first explanation for the difference in the robustness magnitudes across vision models and enc-LLMs. Our findings are also consistent with YOPO \cite{zhang2019you} and TMD \cite{minh2022textual}. Specifically, YOPO highlights the critical role of the first layer in vision models for AT, while TMD uses the last layer to detect AEs in enc-LLMs.

We further leverage our insights on the impact of the manifold ID on robustness and generalization and hypothesize that
\textit{perturbing the intermediate layer $l$ with the highest off-manifold AE ratio (equivalently lowest ID) 
should
achieve high robustness at low computational cost.}
Intuitively, a lower ID corresponds to a higher proportion of OFM-AEs generation and a greater improvement in robustness by the manifold conjecture. Additionally, layers closer to the output result in shorter PGD chains, leading to efficient AT. 
To this end, we propose \system{}\footnote{The code is publicly available at: \url{https://github.com/EnesAltinisik/SMAAT-25/tree/main}}, a Scalable Manifold Aware AT approach that applies AT at the layer with the highest proportion of OFM-AEs. 
We found that this critical layer is consistently the \textbf{last} layer across enc-LLMs on several applications. Hence, for enc-LLMs, \system{} leads to a significant speed-up of AT by avoiding a full backward pass as it calculates the gradients only until the last layer rather than the entire network. Also, empirically, we show that we achieve a higher robustness. 
Yet, this ID trend is different for vision or dec-LLMs, where we find that the first layer is always the one with the lowest ID. 
In these cases, \system{} effectively reduces to standard AT.

\input{figures/fig_overview}

Given enc-LLMs continue to play a crucial role in machine learning pipelines, there is significant value in effectively enhancing their robustness. 
To this end, we evaluated SMAAT in improving the robustness of (1) classifiers, (2) safety filters, and (3) retrievers within Retrieval-Augmented Generation (RAG). \system{} achieved state-of-the-art (SOTA) runtime and robustness results on all tasks, while maintaining clean accuracy (generalization) comparable to standard training. Specifically, in sentiment and content classification tasks, \system{} improved the robustness of enc-LLMs on AGNEWS~\cite{zhang2015character}, IMDB~\cite{maas-etal-2011-learning}, and YELP~\cite{zhang2015character} datasets by 8.6\%, 15.7\%, and 28.8\% for BERT~\citep{devlin-etal-2019-bert} and by 6.0\%, 5.8\%, and 19.0\% for RoBERTa~\citep{liu2019roberta}, respectively. For safety filtering, \system{} accurately identified harmful prompts generated by GCG \cite{zou2023universal}, achieving 97-100\% accuracy. In RAG experiments, \system{} significantly enhanced the robustness of the Contrevier model \cite{izacard2021unsupervised} on the RAG setup, achieving over 80\% robustness against poisoning attacks \cite{zhong2023poisoning}. Besides, \system{} required only about 25-33\% of the GPU time compared to the standard AT.  
A summary of all results is presented in Fig.~\ref{fig:benchmark-all}.

In summary, the major contributions of our work are:
\begin{enumerate}
\item  An explanation for the discrepancy in the robustness and generalization trends in foundation models.
\item \system{}, a novel AT algorithm that leverages the intrinsic dimensionality across layers of foundational models to control robustness, generalization and scalability.
\item Comprehensive experiments demonstrating enhancements in robustness and scalability while keeping accuracy in classification and retrieval tasks compared to standard AT.
\end{enumerate}

%% file: figures/llama_teaser.tex
\begin{figure}[t]%
\centering
\includegraphics[width=1\columnwidth]{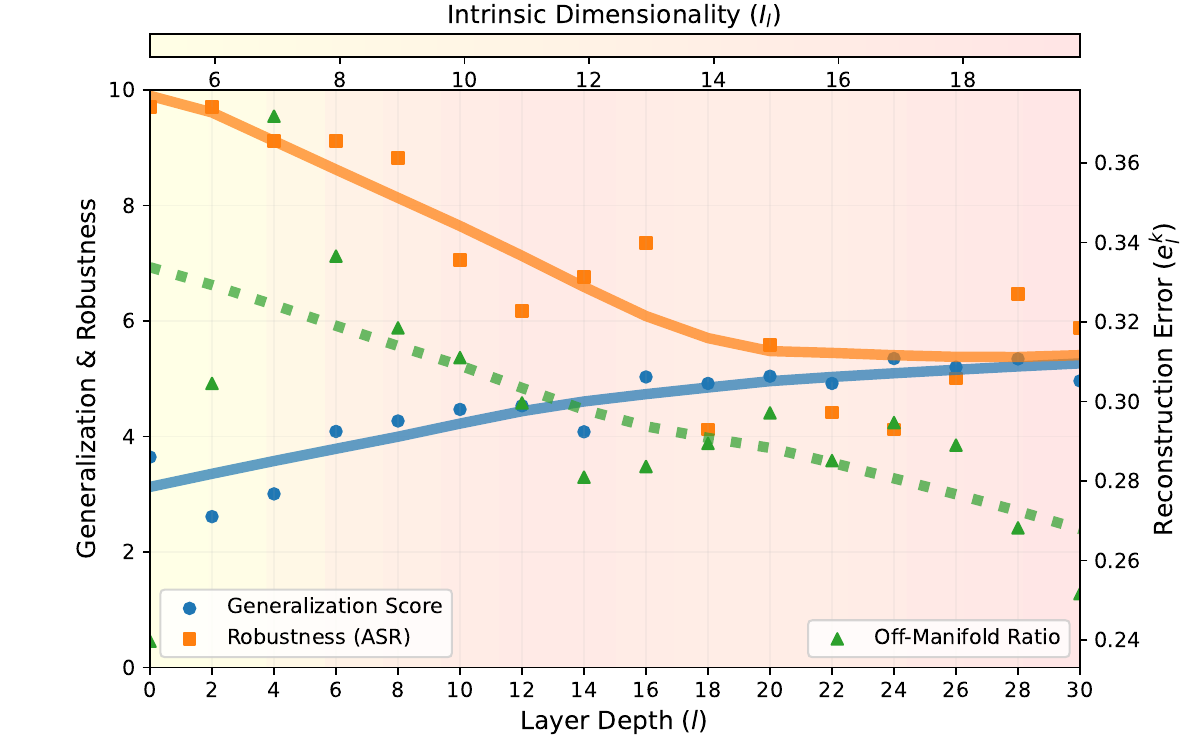}
\caption{Impact of applying LAT at different layers of the LLaMA-2 model, illustrating the relationship between Intrinsic Dimensionality (background color), Generalization (blue), Robustness (orange), and Off-Manifold Ratio (green, based on reconstruction error). 
Markers show the average of measured values across multiple training configurations; lines depict overall trends.
The off-manifold ratio measures the percentage of adversarial examples that fall outside the data manifold using reconstruction error. As we move to deeper layers, the Intrinsic Dimensionality increases, resulting in a decrease in the off-manifold ratio. According to the manifold conjecture, this leads to an increase in generalization (more on-manifold samples) and a decrease in robustness.}

\label{fig:teaser}
\end{figure}

%% file: figures/fig_intor_compare.tex
\begin{figure}[t]%
\centering
\includegraphics[width=1\columnwidth]{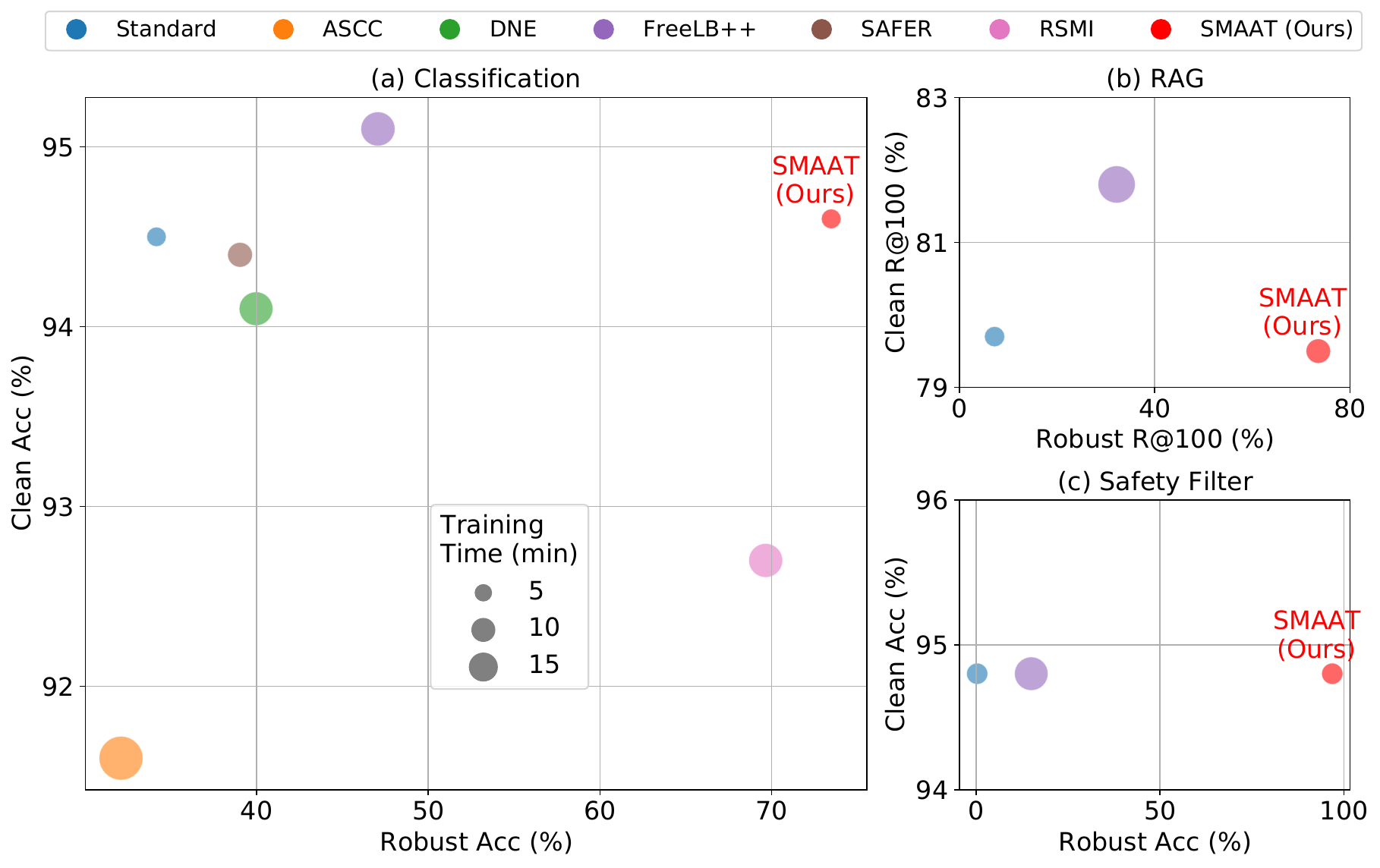}
\caption{Comparison of \system{} robustness (x-axis), generalization (y-axis), and run time (marker size) against baselines for robustifying (a) topic classifiers, (b) retriever models in the Retrieval Augmented Generation (RAG) setup and (c) safety filters for decoder-based LLMs. \system{} significantly enhances model robustness compared to seven different baselines, while maintaining nearly the same clean accuracy. Besides, it is significantly more scalable than AT (marker size).}
\label{fig:benchmark-all}
\end{figure}

%% file: figures/fig_overview.tex
\begin{figure*}[t]
\centering
\includegraphics[width=1\linewidth]{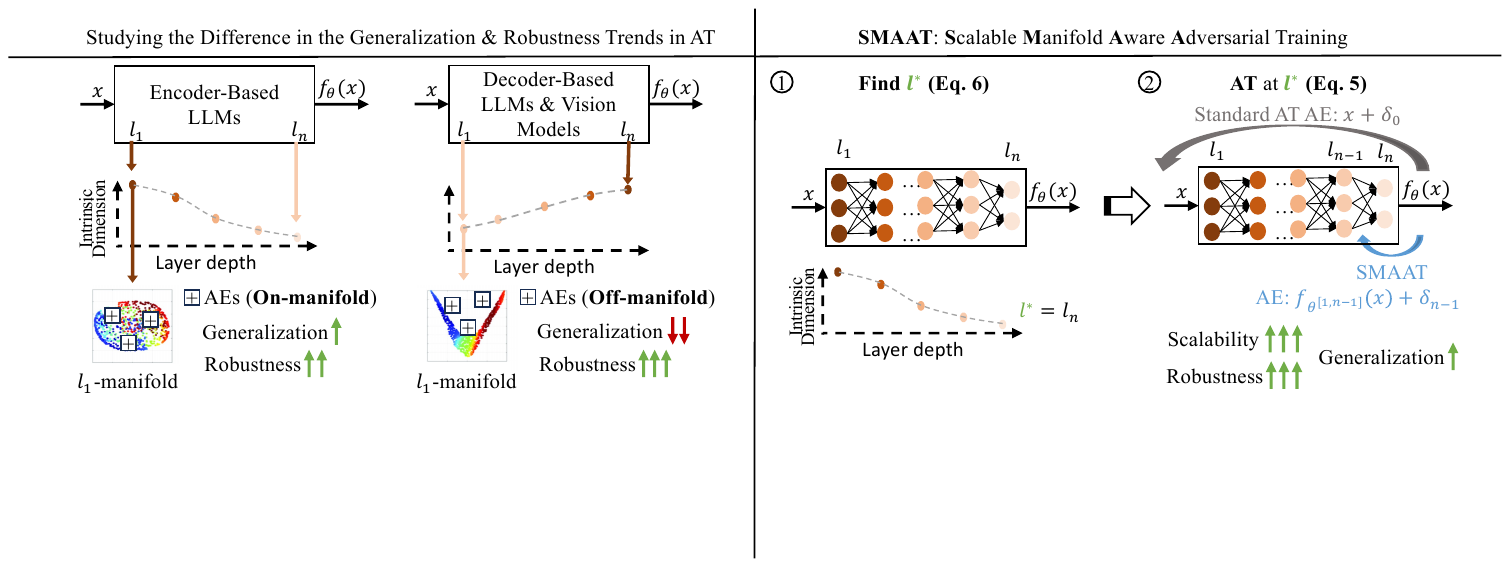}
\caption{%
Left: In classical AT, adversarial examples (AEs) are created in the data layer. For encoder LLMs, the intrinsic dimensionality tends to be high in the initial layers and therefore the AEs created tend to be on-manifold which results in better generalization. In Vision and decoder LLMs, we observe the opposite behavior and AEs tend to be off-manifold resulting in better robustness. Right: The key idea of SMAAT is to create AEs in intermediate layers where the intrinsic dimensionality is low and AEs will tend to be off-manifold. This results in better robustness while (surprisingly) maintaining generalization. The speed-up in SMAAT is due to the fact we need shorter backprop chains to create AEs in intermediate layers.}
\label{fig:approach}
\end{figure*}

%% file: related_work.tex
\section{Related Work}
\label{related_work}
\textbf{Adversarial Training (AT)} aims at robustifying a model against AEs which are imperceptibly perturbed inputs that can lead to incorrect predictions.
Formally, AT seeks optimal parameters \( \theta^{*} \) for a classifier \( f_{\theta}(x) \) that remains robust to perturbations \( \delta \) within a norm ball:  
\begin{equation}
\min_{\theta} \mathbb{E}_{(x,y)\sim\mathcal{D}}\big[ \max_{\|\delta\| \leq \epsilon } \ell(f_{\theta}(x+\delta), y )  \big],
\label{eq:adversarial_loss_}
\end{equation}  
where \( \ell \) is the loss function and \( \mathcal{D} = \{(x,y)\}_{i=1}^{|\mathcal{D}|} \) represents the training data. The outer minimization is typically solved using stochastic gradient descent (SGD), while the inner maximization is addressed via projected gradient descent (PGD)~\citep{madry2017towards}.  
LAT extends AT by applying adversarial perturbations to the model’s latent representations instead of its inputs \cite{casper2024defending,sheshadri2024latent}. Given a model with parameters \( \theta = (\theta_1, \theta_2) \), it computes \( h_{\theta_2} \circ g_{\theta_1} \), where \( g_{\theta_1} \) is a feature extractor that maps inputs to latent representations \( \ell_i = g_{\theta_1}(x_i) \), and \( h_{\theta_2} \) maps these latents to predictions \( \hat{y}_i = h_{\theta_2}(\ell_i) \).  
The standard AT objective under an \( L_p \)-norm constraint of \( \epsilon \) is:  
\begin{align} 
    \min_{\theta} \sum_i \max_{\delta_{i}} \; \mathcal{L}(h_{\theta_2}(g_{\theta_1}(x_{i} + \delta_{i})), y_i) \notag 
\end{align}  
Similar to input-space AT, the inner maximization finds the worst-case perturbation \( \delta_{i} \), while the outer minimization updates \( \theta \), both solved via gradient-based methods. However, AT with \( P \)-step PGD is significantly more computationally expensive than standard training, as it requires \( P \) forward-backward passes per update, compared to just one in standard SGD.

\noindent\textbf{Adversarial Attacks on LLMs.}
Adversarial attacks on LLMs are more challenging than on vision models due to the discrete nature of text and tokenization. 
Early attacks on enc-LLMs used word substitutions guided by embedding similarity \citep{jin2020bert}, synonymity \citep{zang2019word}, or masked language models \citep{li2020bert}, primarily to flip classifier outputs. 
In contrast, recent attacks on dec-LLMs focus on alignment breaking objectives by appending adversarial prefixes or suffixes. 
For example, AutoDAN \cite{liu2023autodan} and PAIR \cite{wei2023jailbroken} craft prompts that encourage harmful completions, while Greedy Coordinate Gradient (GCG) attack~\cite{zou2023universal} appends a gradient based generated adversarial suffix to user inputs.
GCG combines affirmative prompting \citep{wei2023jailbroken,carlini2023are} with greedy and gradient-based discrete optimization \citep{shin2020autoprompt}, and is notable for its strong transferability across prompts and models.

\noindent\textbf{Manifold-Based Defenses.} 
The manifold conjecture stands as one of the most compelling explanations for the susceptibility of deep neural networks to AEs 
\cite{tanay2016boundary,DBLP:conf/iclr/GilmerMFSRWG18,shamir2021dimpled}. The conjecture posits that data resides on a low-dimensional manifold within a high-dimensional representation space and that the deepnet learns to approximate this manifold. Consequently, an off-manifold sample, deviating from this foundational manifold, leads to undefined behavior of the model. 
This conjecture has inspired a novel line of defenses against adversarial attacks on images \cite{samangouei2018defensegan,meng2017magnet,song2017pixeldefend,schott2018towards} and text \citep{minh2022textual}. 
These methods approximate the data manifold and, during testing, project samples onto this manifold to either detect or correctly classify AEs.
Differently, we leverage the manifold conjecture during training to improve both robustness and scalability of AT.

\noindent\textbf{Robustness and Generalization Trends in AT.} 
AT is recognized for enhancing model robustness in both vision and enc-LLMs~\cite{zhang2019theoretically, altinisik2022impact}. 
While this improvement in robustness comes at the cost of increased generalization error in vision models \cite{zhang2019theoretically}, AT enhances generalization in enc-LLMs \cite{altinisik2022impact, villa}. There has been recently efforts relating robustness to OFM-AEs and generalization to ONM-AEs \cite{stutz2019disentangling, xiao2022understanding}. We extend this line of research by connecting these trends to the intrinsic dimensionality of intermediate layers of a network.

\input{figures/fig_es_dimensionality}

\noindent\textbf{Scalable AT.} 
Different optimizations have been proposed to mitigate the cost of the PGD additional forward-backward passes, including  (i) replacing multi-step PGD with a single-step FGSM~\cite{shafahi2019adversarial}; (ii) omitting redundant computations during PGD~\cite{zhang2019you}; 
(iii) combining FGSM with random initialization~\cite{wong2020fast}.
While these approaches alleviate the PGD overhead, they also come with limitations. 
As a solution, FreeLB \cite{Zhu2020FreeLB} proposes accumulating model parameter gradients over multiple batches. 
We achieve scalability through an alternative method, \ie by leveraging the manifold conjecture to perturb intermediate layers and thus reduce the length of backward-forward PGD chains.

\noindent\textbf{ID Estimation.}
It is well known that most real world data lives in a low-dimensional space relative to the ambient space 
where it is defined. The ID of data is the minimum number of variables necessary 
to characterize important properties of the data. Singular Value Decomposition (SVD) is a well known
method to estimate the ID assuming the data lives on a linear manifold~\cite{stewart1993early}. While there have been many proposals to estimate the ID in non-linear settings, we will use the relatively recent twoNN 
ID-estimator based on the observation that for every data point $x$, the ratio $\mu$  of distance of $x$ to it's second and first nearest neighbor follows a Pareto distribution $f(\mu|I) =I\mu^{-(I+1)}$, where $I$ is the ID. It can be estimated as $I = \frac{\log(1 - F(\mu))}{\log(\mu)}$, where $F(\mu)$ is the empirical CDF~\cite{facco2017estimating}.

%% file: figures/fig_es_dimensionality.tex
\begin{figure*}[!t]
\centering
\includegraphics[width=1\textwidth]{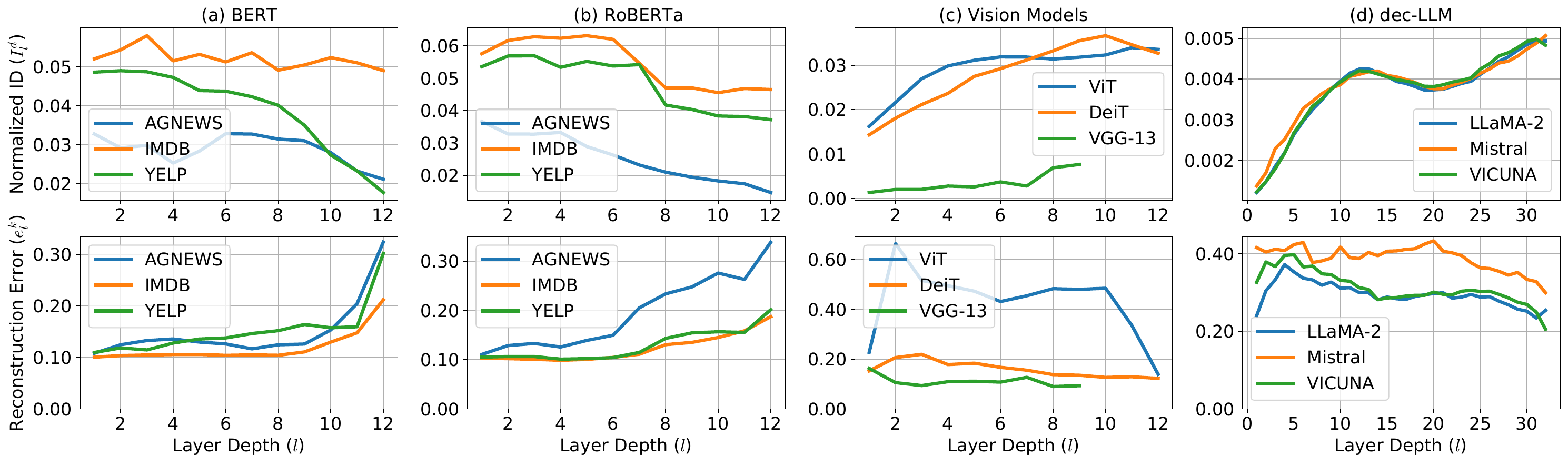}
  \caption{The ID (row 1) trend follows the inverse OFM-/ONM-AEs ratio (row 2) trend. The average projection error ($e_l^k$) is used as a proxy for estimating the OFM-/ONM- AEs ratio. The ID is computed using the twoNN approach. Enc-LLMs (BERT, RoBERTa) have decreasing ID and OFM-AEs proportions trends unlike vision models and dec-LLMs that have increasing ID and ONM-AEs trends.}
    \label{fig:k_effect}
\end{figure*}

%% file: approach.tex
\vspace{-5pt}
\section{Notation}  
Consider a deep neural network classifier \( f_{\theta} \) with \( n \) layers and parameters \( \theta = \{\theta^{(i)}\}_{i=1}^{n} \). We define the transformation spanning layers \( l_i \) to \( l_j \) as \( f_{\theta^{[i,j]}} = f_{\theta^{(j)}} \circ \dots \circ f_{\theta^{(i)}} \), with \( \bar{f}_{\theta^{[i,j]}} \) denoting its standardized version across the dataset \( \mathcal{D} \). The input matrix \( \mathcal{X} \in \mathbb{R}^{d \times |\mathcal{D}|} \) stacks all samples \( \{x_i\}_{i=1}^{|\mathcal{D}|} \), while \( f_{\theta^{[1,l]}}(\mathcal{X}) \in \mathbb{R}^{d_l \times |\mathcal{D}|} \) stacks their transformed representations.  
The \textbf{data manifold} is defined by the high-density region of the sample distribution, distinguishing \textbf{on-manifold} (high-density) from \textbf{off-manifold} (low-density) points. Similarly, the \textbf{$l^\text{th}$ layer manifold} is characterized by the density of standardized representations \( \{\bar{f}_{\theta^{[1,l]}}(x_i)\}_{i=1}^{|\mathcal{D}|} \), approximated via the eigenspace of the covariance matrix \( \bar{f}_{\theta^{[1,l]}}(\mathcal{X}) \bar{f}_{\theta^{[1,l]}}(\mathcal{X})^T \). We denote its eigenvectors and eigenvalues as \( (U_l, \Sigma_l) \), where \( U_l \in \mathbb{R}^{d_l \times d_l} \) and \( \Sigma_l \in \mathbb{R}^{d_l \times d_l} \).  

The \textbf{projection error} on the top \( k \) eigenvectors is:  
\begin{equation}
e_{l}^{k}(x) = \bar{f}_{\theta^{[1,l]}}(x) - U_l^{k}{U_l^{k}}^T \bar{f}_{\theta^{[1,l]}}(x), \quad \forall x \in \mathcal{X}.
\end{equation}  
The \textbf{eigen-based manifold dimension} \( k_l \) is the minimum \( k \) required for a reconstruction error \( \gamma \):  
\begin{equation}
k_l = \min_k \left\{ k \in [1,d_l] \;\Big|\; \sum_{x\in \mathcal{D}} \| e_{l}^{k}(x) \|_2 \leq \gamma \right\}.
\end{equation}  

We classify adversarial examples (AEs) based on projection error:  
- A \textbf{(\(\gamma,l\))-off-manifold AE} (OFM-AE) has \( e_l^k > \gamma \).  
- A \textbf{(\(\gamma,l\))-on-manifold AE} (ONM-AE) has \( e_l^k \leq \gamma \).  
Finally, we denote the \textbf{Intrinsic Dimension (ID)} of layer \( l \), estimated via TwoNN, as \( I_l \), and its \textbf{normalized ID} as \( I_l^d = I_l / d_l \).

\section{Exploring Layerwise ID and Its Effects}
\label{seq:approach}

Our study is motivated by the not-well understood difference in the generalization and robustness trends between adversarially trained vision models and LLMs. 
Specifically, (1) robustness improvements are consistently more pronounced in vision models~\cite{madry2017towards,altinisik2022impact}, (2) generalization either remains stable or improves in enc-LLMs~\cite{tsipras2018robustness,Zhu2020FreeLB}, whereas it deteriorates in vision models, and (3) dec-LLMs achieve the optimal trade-off when AT is applied across multiple network layers \cite{casper2024defending,sheshadri2024latent}.

We demonstrate that the varying trends observed across different network architectures are closely linked to the layer-wise distribution of OFM-AEs.
Notably, we find that in the first layer, where conventional AT is applied, adversarial examples for enc-LLMs tend to be more on-manifold, while vision models and dec-LLMs show a higher proportion of off-manifold examples (see \figref{fig:approach}, \textit{left}).
This explains the generalization and robustness trends as, by the manifold conjecture, higher proportions of ONM and OFM AEs lead, respectively, to better generalization and robustness~\cite{ma2018characterizing,Stutz_2019_CVPR,alemany2020dilemma,Li_2021_ICCV}.

\input{figures/mid_layer}
\textbf{ID \& ONM/OFM Relationship} 
We hypothesize that lower intrinsic dimensionality ($I_l^d$) leads to a higher proportion of OFM-AEs, whereas higher $I_l^d$ yields more ONM-AEs. 
Formally, let $\mathcal{M} \subset \mathbb{R}^n$ be a smooth, compact, low-dimensional manifold with intrinsic dimension $I_l^d \ll n$, embedded in $\mathbb{R}^n$.
Let $f: \mathbb{R}^n \to \mathbb{R}^k$ be a classifier trained on data sampled from $\mathcal{M}$.
Since the classifier learns $p(y \mid x)$ without access to the true generative distribution $p(x)$, it lacks explicit knowledge of the underlying data manifold $\mathcal{M}$. 
Consequently, the loss gradient $\nabla_{\delta} \mathcal{L}$ often contains components that are orthogonal to the tangent space of $\mathcal{M}$. 
As a result, the perturbation $\delta$ is unlikely to remain entirely within the manifold and typically includes a non-zero component in the orthogonal direction.
Moreover, when $I_l^d$ is small relative to $n$, the proportion of $\delta$ that lies off the manifold increases. 
Thus, adversarial examples generated via gradient-based methods are more likely to fall outside the data manifold, i.e., to be off-manifold adversarial examples.
 
We base this conclusion on an empirical investigation into the relationship between the ID and the proportion of ONM/OFM AEs across different layers of various deep neural network models, including two enc-LLMs (BERT and RoBERTa), three vision models (ViT~\cite{dosovitskiy2020image}, DeiT~\cite{touvron2021training}, VGG-13~\cite{simonyan2014very}), and three dec-LLMs (LLaMA-2-7b-chat~\cite{touvron2023llama}, Vicuna~\cite{zheng2024judging}, Mistral-7b-instruct~\cite{jiang2023mistral}) on several datasets (refer to Appendix \ref{sec:relation} for details). Results are reported in Figure \ref{fig:k_effect}. The ID ($I_l^d$) in row 1, is computed using twoNN estimation\cite{facco2017estimating} and follows the OFM-AEs trend (row 2). To assess the proportion of OFM/ONM AEs across different layers, we use the reconstruction error $e^k_l$ at layer $l$ (row 2) as a metric, \ie higher projection error corresponds to higher proportion of OFM-AEs.

\textbf{ID \& Robustness/Generalization Relationship}

We extend our evaluation of the relationship between intrinsic dimensionality (ID), robustness, and generalization by applying adversarial training (AT) to different layers of dec-LLMs (LLaMA-2-7B-chat), enc-LLMs (BERT), and vision models (VGG).
For the dec-LLM, we train LLaMA-2-7B-chat on the LAT dataset~\cite{sheshadri2024latent} using various hyperparameter configurations (see Appendix~\ref{sec:app:lat} for details), applying LAT at every even-numbered layer. 
To evaluate generalization, we use MT-Bench~\cite{zheng2023judging}, while robustness is assessed via failure rates against the GCG attack , which generates adversarial suffixes that bypass language model safety alignment. 
Additional results using the PAIR~\cite{chao2023jailbreaking} attack are provided in Appendix~\ref{sec:app:lat}.
As shown in Fig.~\ref{fig:mid_layer}(a), applying LAT to lower layers (light blue markers) enhances robustness but reduces generalization (bottom-right), while applying it to upper layers (dark blue markers) improves generalization at the cost of robustness (top-left). This trend aligns with the intrinsic dimension (ID), which decreases in higher layers, as well as with the distribution of ONM/OFM adversarial examples. These results also explain the superiority of the LAT approach of \citealp{sheshadri2024latent} which applies AT to four evenly spaced layers across the network.

We conducted a similar experiment on the BERT model using the YELP dataset (see Sec.~\ref{sec:sent} for details). Our results empirically show that, following the decreasing ID trend, applying AT at higher layers leads to increased robustness as the layer index $i$ increases.
Figure~\ref{fig:mid_layer}(b) presents the results of AT on BERT for the YELP dataset. As expected, robustness (measured under the TextFooler attack) improves with higher layer indices while generalization remains unaffected. 

We also extend this analysis to vision models. We selected VGG-13 and trained it on CIFAR-10~\cite{cifar10} using AT applied to every ReLU layer over 20 epochs. We varied the attack strength ($\epsilon = 0.031$–$0.2$) and learning rate ($lr = 0.01$–$0.001$), evaluating robustness via RobustBench~\cite{}. As shown in Fig. ~\ref{fig:mid_layer}(c), adversarial training on lower layers (light blue) improves robustness but reduces generalization, whereas training on higher layers (dark blue) yields the opposite pattern. These results suggest that vision models resemble dec-LLMs in terms of their generalization–robustness tradeoff.
Finally, our findings are consistent with those shown in Fig. ~\ref{fig:k_effect}, where a higher ratio of OFM-AEs in deeper layers correlates with improved robustness in both enc-LLMs and vision models.

Beyond providing an explanation for the difference in robustness and generalization trends in vision models, enc- and dec-LLMs, results of \figref{fig:k_effect} shed light on an interesting trend for the ID across layers. While the ID is increasing in vision and dec-LLMs, it is monotonically decreasing in enc-LLMs (Bert and RoBERTa). This motivated the design of a new algorithm that controls the gains in robustness and generalization via controlling the proportions of ONM- and OFM-AEs. This can be achieved by generating AEs in the intermediate layers of the deepnet based on their IDs. To this end, we propose \system{}\ (See \figref{fig:approach}, \textit{right}), an AT algorithm that aims at achieving high robustness by perturbing the layer $l^*$ with the lowest ID to generate a high proportion of OFM-AEs. A side effect of perturbing an intermediate layer as opposed to the input one is significantly improving scalability by a factor $\cO\Big(P(n-l^{*}) \Big(\max\!\Big(\{d_{l_i} | l_i \in [1, l^{*}-1]  \} \Big)-d_{l^*} \Big) \Big)$ as it results in shorter $P$-step PGD chains.

\input{Tables/tab_results_nlp}

\section{\textbf{SMAAT}: \textbf{S}calable \textbf{M}anifold \textbf{A}ware \textbf{A}dversarial \textbf{T}raining}
\label{sec:smaat}
\vspace{-2pt}
\noindent We propose \system{} an AT algorithm that aims at improving the robustness and scalability of standard AT by generating AEs in the layer leading to the highest proportion of OFM instead of from the input layer as classically.
Specifically, given a pretrained model $f_{\theta}$, \system{} intentionally generates a higher proportion of OFM-AEs to enhance robustness.
Formally, \system{} solves the augmented AT objective:
\begin{equation}\label{eq:adv_loss_8}
\begin{aligned}
\min_{\theta} \quad & \mathbb{E}_{(x,y)\sim\mathcal{D}}\left[ \max_{\|\delta\| \leq \epsilon } \ell(f_{\theta}(x+\delta), y )  \right]\\
\textrm{s.t.} \quad & (x+\delta) \text{ is $(\gamma,1)$-OFM}. \\
\end{aligned}
\vspace{-1pt}
\end{equation}
While the manifold conjecture refers to input space AEs, we relax it to encompass intermediate layers as well, \ie we chose the perturbation $\delta$ that results in an OFM-AE at any layer across the deepnet. Intuitively, AEs that are off- or on- the transformed data manifold in any layer also affect robustness. The relaxed objective becomes:    
\begin{equation}\label{eq:at_obj}
\begin{aligned}
\min_{\theta} \quad & \mathbb{E}_{(x,y)\sim\mathcal{D}}\big[ \max_{\|\delta\| \leq \epsilon } \ell(f_{\theta}(x+\delta), y )  \big]\\
\textrm{s.t.} \quad & \exists l\!\in\![1,L]\!:\!(x+\delta) \text{ is } (\gamma,l)\text{-OFM. } \\
\end{aligned}
\end{equation}
Note that solving the above objective using the method of Lagrangian Multipliers is possible, but it would require an approximation of the manifold to characterize OFM-AEs. Such approximations are either computationally expensive (\eg GAN \cite{xiao2018generating}) or overly simplistic (\eg eigenbasis \cite{xiao2022understanding}). \system{} 
uses 
an alternative approach to find OFM-AEs by perturbing the layer with the lowest ID without the need for manifold approximations. 
Particularly, \system{} applies AT at $l^{*}$-th layer where the layer with more OFM-AEs composition (lowest ID):
\begin{equation}
\min_{\theta}  \mathbb{E}_{(x,y)\sim\mathcal{D}}\Big[ \max_{ \|\delta_{l^*}\| \leq \epsilon_{l^*} }\ell\Big(f_{\theta^{[l^*,n]}}\Big(f_{\theta^{[1,l^*]}}(x)+\delta_{l^*}\Big), y \Big)  \Big].
\label{eq:adversarial_loss_approach}
\end{equation}
We additionally choose $l^*$ to correspond to the layer with the highest index as this would lead to better scalability (shorter PGD chains).
as well as to potentially the highest proportion of off-manifold AEs at any intermediate layer, 
\ie
\begin{equation}
\label{eq:opt_layer_fine}
l^* =  \max_{l} \Big\{ l \in [1,n] \Big|  I_l^* \leq I_{i} \forall i<l   \Big\}.
\end{equation}

\input{Tables/tab_text_runtime}

\noindent\textbf{Complexity.}
Computing the layers IDs and searching for the optimal layer $l^{*}$ to perturb are done once per model and per task. Thus, they incur marginal overhead.
For the AT part, when a $P$-step PGD attack is used, this results in $P$ forward-backward passes with length $(n-l^{*}+1)$ instead of $n$. The run-time of every forward/backward pass depends on the layer dimensionality, \ie $O(d_{l-1} \times d_l)$ for the $l^{\text{th}}$ layer. Overall, the complexity of one \system{} forward-backward is $\cO( (n-l^*+1)\max_{l \in [l^*,n]}(d_l)^2 ))$. As a result, \system{} is more efficient than standard AT by a factor of $\cO( P \times l^{*} \cdot ( \max_{i\in [1,n]}{(d_i)^2} - \max_{j\in [l^*,n]}{(d_j)^2} ))$. In the case Enc-LLMs (\eg BERT and RoBERTa), $l^{*}$ is equal to $n$. The total run-time can be simplified to $\cO( P \cdot (d_n)^2 )$ where $d_n$ is the number of classes. Typical enc-LLMs tasks consist of less than five classes~\cite{wang2019glue} which makes the factor of classes small enough to be negligible. Hence, \system{} enhances the efficiency of the AEs generation process by a factor of $l \cdot O(\max(d_l))$. This improvement practically eliminates the cost of the AE generation process.

%% file: figures/mid_layer.tex
\begin{figure*}
  \begin{center}
    \begin{minipage}[a]{0.32\textwidth}
    \centering
    \includegraphics[width=1\columnwidth]{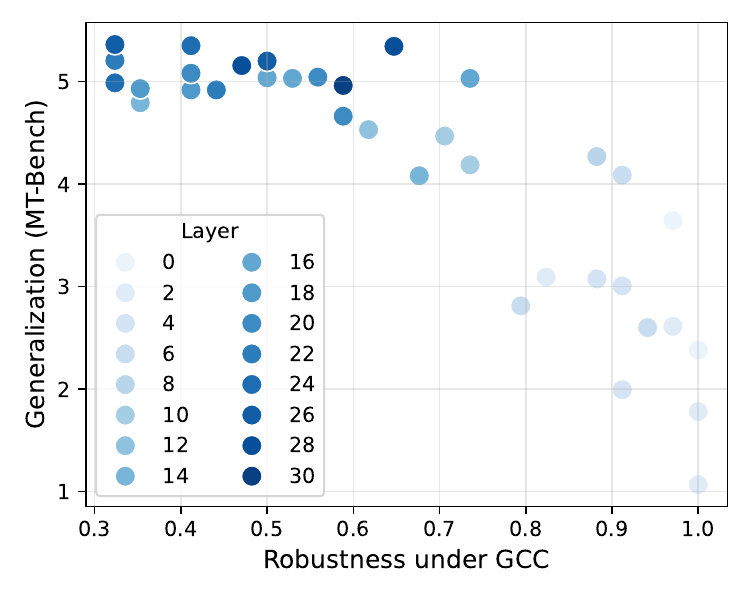}
    \centerline{(a) LLaMA-2 }
  \end{minipage}
  \begin{minipage}[a]{0.32\textwidth}
    \centering
    \includegraphics[width=1\columnwidth]{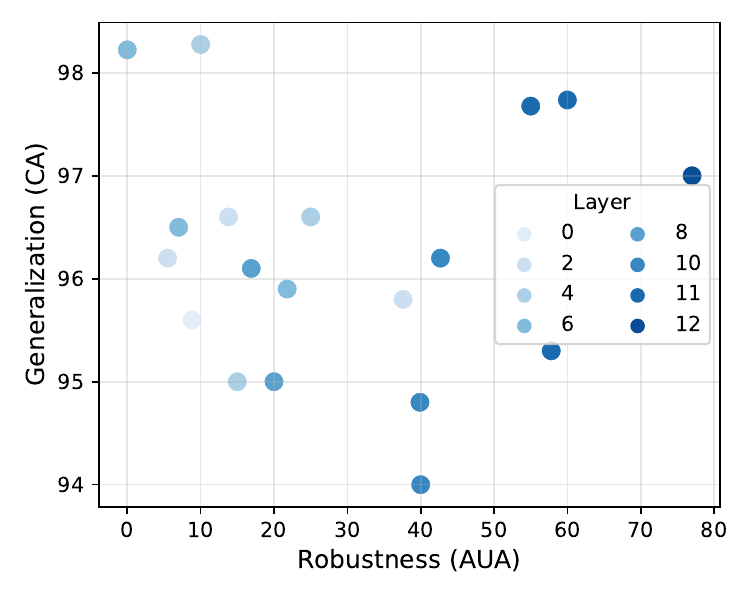}
    \centerline{(b) BERT}
  \end{minipage}
  \begin{minipage}[a]{0.32\textwidth}
    \centering
    \includegraphics[width=1\columnwidth]{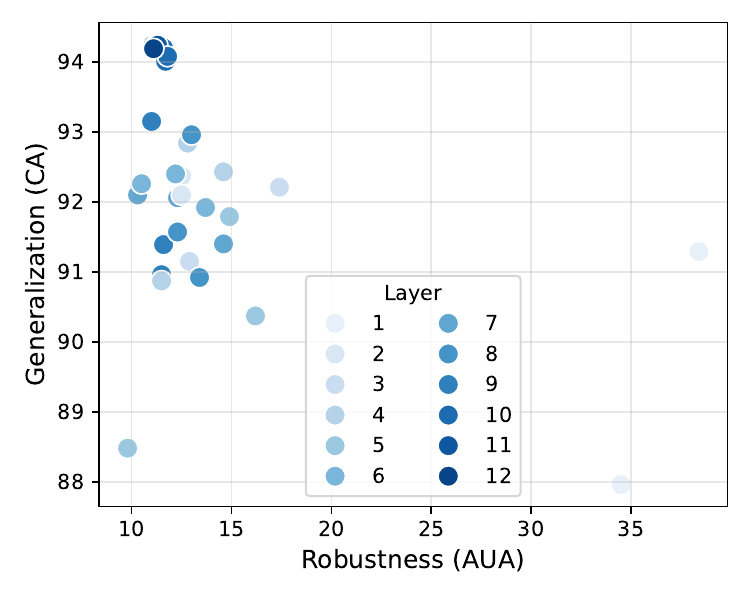}
    \centerline{(c) VGG}
  \end{minipage}
\caption{Layer-wise effects of adversarial training on robustness and generalization across model types.
Each subplot shows the impact of applying AT at different layers of (a) LLaMA-2 (dec-LLM), (b) BERT (enc-LLM), and (c) VGG (vision model). Marker colors transition from light blue (lower layers) to dark blue (higher layers). Observed trends align with changes in intrinsic dimensionality and the distribution of on- vs. off-manifold adversarial examples, as shown in Figure~\ref{fig:k_effect}.}
  \label{fig:mid_layer}
  \end{center}
\end{figure*}

%% file: Tables/tab_results_nlp.tex
\begin{table*}[t]
\caption{Robustifying sentiment classifiers on AGNEWS, IMDB, and YELP datasets. We report the clean accuracy (CA) and the robust accuracy under attack (AUA) with PWWS (PW), TextFooler (TF), and Bert-Attack (BA), along with the average robust accuracy (AR) across these three attacks.
The best performance is \textbf{bold}ed. Across all data sets, \system{} achieves high robustness (AR) while maintaining high generalization (CA).} %
\centering
\fontsize{9}{11}\selectfont
\setlength{\tabcolsep}{1.7mm}
\begin{tabular}{c|c|c|cccc|c|cccc|c|cccc}
\toprule 
 \multirow{3}{*}{Model}  & \multirow{3}{*}{Defense}     & \multicolumn{5}{c|}{AGNEWS}                                            & \multicolumn{5}{c|}{IMDB}        & \multicolumn{5}{c}{YELP}          \\ %
                                & & \multirow{2}{*}{CA}    & \multicolumn{4}{c|}{AUA} & \multirow{2}{*}{CA}    & \multicolumn{4}{c|}{AUA} & \multirow{2}{*}{CA}    & \multicolumn{4}{c}{AUA} \\ 
                                & & & PW                & TF                & BA           & AR  &                 & PW                & TF                & BA             & AR    &                 & PW                & TF                & BA       & AR          \\ 
    \midrule
    \multirow{8}{*}{\rotatebox[origin=c]{90}{BERT}}  &Standard                      & 94.5         & 36.9              & 28.1              & 37.5          & 34.2 & 92.2              & 15.0              & 5.8               & 5.4               & 8.7 & \textbf{97.0}     & 12.2              & 6.5               & 5.3                & 8.0 \\
    \cmidrule{2-17}
    &ASCC                         & 91.6         & 32.8              & 31.4              & 32.1          & 32.1 & 88.5              & 15.1              & 12.4              & 11.2              & 12.9 & 91.5              & 19.4              & 15.7              & 12.2               & 15.8 \\
    &FreeLB++             & \textbf{95.1}        & 47.9          & 51.5            & 41.8          & 47.1 & \textbf{93.2}     & 12.5         & 45.3              & 39.9  & 32.6 & 95.6& 19.3        & 8.8           & 3.7            & 10.6 \\
    \cmidrule{2-17}
    &SAFER                        & 94.4         & 39.3              & 35.5              & 42.3          & 39.0 & 92.3  & 41.4              & 39.1              & 30.7              & 37.1 & 95.4              & 29.8              & 25.8              & 23.7       & 26.4 \\
    &TMD                          & 94.3         & 70.0              & 50.0              &55.2& 58.4 & 92.2             & 38.7              & 44.2              & 33.7              & 38.9 & 95.2              & 36.8              & 40.9              & 28.6                   & 35.4 \\
    &RSMI                         & 92.7         & \textbf{76.1}  & 63.2  & NA\fnm       & NA & 92.2              & 58.7  & 56.4  &  NA\fnmNon         & NA & 95.4    &45.3& 52.3  &  NA\fnmNon     & NA \\
    \cmidrule{2-17}    
    \cmidrule{2-17}
    &\textbf{\system{} (Ours)}     & 94.6      & 73.5              & \textbf{72.2}              & \textbf{74.7}          & 73.5 & 92.2              & \textbf{63.6}              & \textbf{77.9}     & \textbf{60.8}              & 67.4 & \textbf{97.0}     & \textbf{77.1}     & \textbf{77.9}              & \textbf{72.8}                   & 75.9 \\
    
    \midrule      
                         
    \multirow{8}{*}{\rotatebox[origin=c]{90}{RoBERTa}}    &Standard       &  94.7  &  30.6     &    23.9   &   37.1  & 30.5 &   94.0     & 8.7    & 2.1  &   0.6   & 3.8 & 97.9 &   23.1    &  14.9     &    9.0     & 15.7 \\
    \cmidrule{2-17}
    &ASCC       & 92.6   & 48.1          & 41.0          & 49.1          & 46.1 & 92.6          & 23.1          & 13.5          & 11.8  & 16.1 & 95.4     & 15.0       & 8.6          & 4.5 & 9.4 \\
    &FreeLB++   & \textbf{95.6}    &  61.0  &  49.8    &  56.6   &   55.8   &  \textbf{94.3}       &  33.6     &    14.6     &   6.1 &  18.1  & 97.0 &  38.6     &  46.0     & 35.2  &39.9   \\ 
    \cmidrule{2-17}
    &SAFER         &  94.6  &   68.9    & 49.3   &  46.1  & 54.8 &  93.9     &  52.8    &   47.1     &  40.6     & 46.8 &   96.6  &  65.6      &  67.9      &   48.3       & 60.6 \\
    &TMD           & 95.0    & 68.3 & 54.0 & 56.7 & 59.7 & 93.3          & 60.5 & 66.8 & 51.6 & 59.6 & 96.6          & 68.9 & 70.9 & 51.0    & 63.6 \\
    &RSMI               & 94.3   &   \textbf{81.9}     &  74.1      &    NA\fnmNon  & NA &  93.0       &  76.2    &   73.4     & NA\fnmNon  & NA & 96.3  &   68.9    &  65.9     &      NA\fnmNon    & NA\\
    \cmidrule{2-17}    
    \cmidrule{2-17}
    &\textbf{\system{} (Ours)}       &   94.6 &  75.6    & \textbf{75.1}       & \textbf{79.9}    & 76.9 &  93.5     & \textbf{77.1}        &   \textbf{78.5}     & \textbf{63.2}    & 72.9 &   \textbf{98.0}  & \textbf{85.4}   &   \textbf{86.4}    &    \textbf{76.0}    & 82.6 \\
    \bottomrule
\end{tabular}%

\label{tab:nlp_res}
\end{table*}

%% file: Tables/tab_text_runtime.tex
\begin{table*}[!ht]
    \centering
    \caption{Run-time results on IMDB dataset. Mean and standard deviation are computed over ten runs. Compared to FreeLB++, training time of \system{}  is lower  by a factor of nearly three.
    }
    \begin{tabular}{ccccccc}
    \toprule
                                                                    & Standard        & ASCC                       & FreeLB++         &  SAFER          & RSMI              & \textbf{\system{}}     \\ \midrule
    \begin{tabular}[c]{@{}c@{}}Training (min/epoch)\end{tabular}    & 5.1 \textpm 0.1& 25.7 \textpm 0.3  & 15.6 \textpm 0.5 & 8.2 \textpm 0.6 & 15.4 \textpm 0.3 & 5.2 \textpm 0.2 \\
    \begin{tabular}[c]{@{}c@{}}Inference (msec/sample) \end{tabular}& 2.4 \textpm 0.1& 41.4 \textpm 0.2   & 2.4 \textpm 0.0  & 2.4 \textpm 0.0 & 5.6 \textpm 0.4   & 2.4 \textpm 0.1\\
     
    \bottomrule
    \end{tabular}
    \label{table:runTime}
\end{table*}

%% file: experiments.tex
\section{Experiments}
\label{experiments}
We assess robustbness, generalization and scalability of three types of models trained with \system{}: (i) classifiers; (ii) retrievers in RAG settings and (iii) safeguarding models employed for moderating content produced by generative models. We consider four different attacks: (i) word substitution \cite{ren-etal-2019-generating,Jin_Jin_Zhou_Szolovits_2020,li-etal-2020-bert-attack}, (ii) adaptive \cite{tramer2017ensemble}, (iii) Greedy Coordinate Gradient (GCG)  \cite{zou2023universal}, and (iv) corpus poisoning  \cite{zhong2023poisoning} attacks. We include a run-time analysis to demonstrate scalability of \system{}.

\noindent\subsection{Sentiment and Topic Classifiers.}\label{sec:sent} We adversarially trained two \textbf{base models}, BERT-base-cased and RoBERTa-base-cased on the \textbf{tasks} of sentiment classification (IMDB~\cite{maas-etal-2011-learning} and YELP~\cite{zhang2015character}) and topic classification (AGNEWS~\cite{zhang2015character}) under \textbf{three input space attacks} including PWWS \cite{ren-etal-2019-generating} (synonym based), TextFooler \cite{Jin_Jin_Zhou_Szolovits_2020} (neighbor based), and BERT-Attack \cite{li-etal-2020-bert-attack} (masked-LM based). Attacks are conducted using the TextAttack framework \cite{morris-etal-2020-textattack} and following the settings introduced by \cite{li-etal-2021-searching}.
We compare \system{} to standard (non-adversarial) training, and six \textbf{baselines} from three families of defenses: (1) Input space AT (ASCC \cite{DBLP:conf/iclr/DongLJ021}), (2) Embedding space AT (FreeLB++ \cite{li-etal-2021-searching}), and (3) Certified defenses (SAFER \cite{ye-etal-2020-safer}, TMD \cite{minh2022textual}, RSMI \cite{minh2022textual}). FreeLB++ focuses on scalability and robustness by minimizing the number of PGD steps and applying AT in the initial layer. TMD leverages manifold features by projecting samples back to the manifold in the last layer. 
\noindent\textbf{Implementation details:} In both base models, $l^{*}=n$, \ie we generate AEs in the last layer. Specifically, we perturb the [CLS] embeddings before the classifier layer by freezing all layers before $l^{*}$. 
Further details are presented in Appendix \ref{sec:app:exp}.

\input{Tables/gcg}

\textbf{Robustness and generalization:} In Table~\ref{tab:nlp_res}, we report results on robustness and generalization. 
On average, \system{} demonstrates superior robustness over all datasets, with an improvement of 8.6\%, 15.7\%, and 28.8\% over the best score for the BERT model, and 6.0\%, 5.8\%, and 19.0\% for the RoBERTa model on the AGNEWS, IMDB, and YELP datasets, respectively. Note that FreeLB++, which perturbs the first layer, showed the best generalization in four out of six cases, as it produces mostly on-manifold examples (the first layer has a high dimension, as illustrated in \figref{fig:k_effect}) which is in line with the manifold conjecture. \system{} maintains generalization in five out of six cases and only shows $0.5$ drop in performance in the case of RoBERTa with the IMDB dataset. 
Besides, \system{} consistently outperforms TMD, even though it is a manifold-based method differently from \system{}, TMD estimates the manifold and projects the input samples onto it before classification witch could lead to inaccuracies in the manifold estimation. \system{} on the other hand is more robust to such inaccuracies as it only leverages the ID for selecting the layer to perturb. 
RSMI shows strong robustness against PWWS, as its masked inference mechanism inherently mimics synonym-based defenses without needing an explicit synonym set \cite{minh2022textual}.
\fnt{RSMI takes about 2k times longer than TextFooler to generate a AE with BERT-Attack, making it unfeasible to test.}

\textbf{Runtime efficiency:} In Table \ref{table:runTime}, we provide details on the training time per epoch and inference time per instance for the IMDB dataset using BERT (RoBERTa has the same architecture). \system{} has comparable efficiency to standard training and is on average 3 times faster than standard AT during training. This is attributed to shorter backpropagation chains as AT in \system{} is performed in the last layer ($l^* =n$). Note that FreeLB++, which injects noise in the first layer, remains inefficient even when PGD is replaced with FGSM \citep{Zhu2020FreeLB, li-etal-2021-searching}, as a full-depth backpropagation chain is still required. Certified defense baselines (SAFER, RSMI) and input space attack (ASCC) are also time-consuming as they either require mapping samples into the manifold or performing an extensive search over word substitutes.
Additional evaluation of \system{} on \textbf{language understanding benchmarks}, including GLUE and advGLUE are reported in Appendix \ref{sec:app:glue}.

\input{Tables/RAG}

\subsection{Safety Filters}
The broad attack surface of LLMs has compelled model owners to implement solutions that extend beyond the safety alignment of a model. Notably, this includes content moderation filters that verify the harmlessness of a model's inputs and outputs \cite{kumar2023certifying, cao2023defending,SGM}. These filters are essentially text classifiers, often based on lightweight enc-LLMs to minimize overhead and preserve the high responsiveness of dec-LLMs \cite{kumar2023certifying}. The robustness of safety filters is also a concern, as they too can be vulnerable to attacks. \textbf{Base models:} We evaluate \system{}'s effectiveness on BERT, RoBERTa, and DistilBERT when used as safety filters. \textbf{Datasets:} The AdvBench dataset \cite{zou2023universal} and the Helpfulness-Harmfulness dataset (HH-RLHF) \cite{bai2022training}.
\textbf{Adversarial attack:}, we use GCG attack. \textbf{Metrics:} we assess generalization accuracy (ACC) for safe and harmful prompts, and robust accuracy under attack (AUA) by measuring the detection rate of harmful prompts augmented with adversarial suffixes. Further details are presented in Appendix \ref{sec:app:gcg}. \textbf{Robustness and generalization:} Results in Table \ref{tab:gcg} show that \system{} significantly enhances the robustness of safety filters against the GCG attack across all models, while maintaining generalization compared to standard training. Also, FreeLB++ improves generalization as it applies AT in the first layer.

\subsection{Retriever Models of RAG}

RAG combines a retriever model, which identifies relevant passages from a large corpus, with a generator model that constructs answers based on the retrieved information. \textbf{Base model:} We use the Contriever model \cite{izacard2021unsupervised}, fine-tuned on the Natural Questions (NQ) \cite{kwiatkowski2019natural} dataset, as our retriever. \textbf{Baselines:} We evaluate the robustness of retriever models within the RAG framework under standard, FreeLB++ and \system{}. \textbf{Adversarial attacks:}, we use poisoning attacks \cite{zhong2023poisoning}, which manipulate the retrieval corpus by generating adversarial passages. As \textbf{metrics}, we use robust recall (RR) measured by how many samples are selected without adversarial passages in the top-k passages, with R@10 and R@100 corresponding to the top-10 and top-100 passages, respectively. \textbf{Robustness and generalization:} Table \ref{tab:rag} shows that the standard model offers limited robustness against the attack, with a significant number of adversarial passages being retrieved. FreeLB++ shows some improvement, reducing the number of adversarial passages retrieved. However, SMAAT significantly enhances robustness, demonstrating a dramatic reduction in the retrieval of adversarial passages compared to the others. 
In terms of generalization, FreeLB++ yields the best results as it applies AT in the first layer (resulting in more ONM-AEs), while both \system{} and standard training exhibit similar performance.

%% file: Tables/gcg.tex
\begin{table}[!t]
    \centering
    \caption{Robustifying dec-LLMs with encoder-based (BERT etc.) safety filters on AdvBench and HH\_RLHF datasets. We report clean accuracy (CA), and robust accuracy under attack (AUA) which indicates the percentage of harmful prompts augmented with adversarial suffices accurately classified as harmful. \textbf{Notice that \system{} achieves very high robustness while other models sometimes completely fail}.}
    \fontsize{9}{11}\selectfont
    \setlength{\tabcolsep}{1mm}
    \begin{tabular}{c|c|cc|cc|cc}
    \toprule 
     \multirow{2}{*}{Dataset} & \multirow{2}{*}{Model}   & \multicolumn{2}{c|}{BERT}               & \multicolumn{2}{c}{RoBERTa}    & \multicolumn{2}{c}{DistilBERT}       \\ %
                            & & CA & AUA & CA & AUA &  CA & AUA          \\  
      \midrule
      \multirow{2}{*}{AdvBench} &Standard              & \textbf{100}  &  0  & \textbf{100}  & 0 & \textbf{99.6} &  0    \\ 
                          &FreeLB++             & 99.6  &  0  & \textbf{100}  & 0 & \textbf{99.6} &  0       \\ 
                           &\textbf{\system{}}            & \textbf{100} & \textbf{100}  &  \textbf{100} & \textbf{99.2} & 99.2 &  \textbf{97.5}     \\ 
      \midrule
       \multirow{2}{*}{HH-RLHF} &Standard              & 95.4 & 26.4 & 94.8  & 0.3 & \textbf{98.7} &  6.5   \\ 
                           &FreeLB++            & \textbf{98.6} & 48.5  & \textbf{98.6}  & 34.0 & 98.5 &  34.1     \\ 
                        &\textbf{\system{}}            & 95.4 & \textbf{51.6}  & 94.8  & \textbf{96.8} & 98.5 & \textbf{39.5}      \\
     \bottomrule
                                                                                                 
    \end{tabular}%
    \label{tab:gcg}
\end{table}

%% file: Tables/RAG.tex
\begin{table}[t]
    \centering
    \caption{Robustifying retriever models of RAG on the Natural Questions dataset. We report Recall@10 (R@10) and Recall@100 (R@100) on the clean corpus for generalization. Robust recall (RR) is measured by how many samples are selected without adversarial passages in the top-k passages, with R@10 and R@100 corresponding to the top-10 and top-100 passages, respectively. During attacks, 10 and 50 adversarial passages are created, denoted as (N=10) and (N=50), respectively. \system{} makes RAG substantially more robust against selecting adversarial passages.}
    \setlength{\tabcolsep}{1mm}
    \fontsize{9}{11}\selectfont
    \begin{tabular}{c|cc|cc|cc}
    \toprule 
      \multirow{2}{*}{Model}   & \multicolumn{2}{c|}{Generalization}               & \multicolumn{2}{c|}{RR (N=10)  }  & \multicolumn{2}{c}{RR (N=50)}       \\ %
      & R@10 & R@100 & R@10 & R@100 & R@10 & R@100         \\  
      \midrule
      Standard & 36.4 & 79.7 & 46.0 & 22.2 & 26.1 & 7.2 \\
      FreeLB++ & \textbf{39.6} & \textbf{81.8} & 75.2 & 60.7 & 51.7 & 32.2 \\
      \textbf{\system} & 34.9 & 79.5 & \textbf{99.5} & \textbf{97.1} & \textbf{85.7} & \textbf{73.5} \\
     \bottomrule
    \end{tabular} 
    
    \label{tab:rag}
             
\end{table}

%% file: conclusion.tex
\section{Conclusion}
\label{conclusion}

In this paper, we explain the varying effects of AT across architectures using the manifold conjecture, showing that ID influences robustness and generalization. Vision and decoder-based models favor off-manifold AEs in early layers, enhancing robustness but harming generalization, while encoder-based LLMs favor on-manifold AEs, preserving generalization with limited robustness gains.  
Leveraging this insight, we introduce \system{}, which improves AT scalability by perturbing the layer with the lowest ID. This reduces training overhead by 25–33\% while boosting robustness across sentiment classification, safety filtering, and retrieval tasks, maintaining generalization.
We plan to explore joint optimization of generalization and robustness by controlling the on/off-manifold AE ratio through perturbations in intermediate layers.

%% file: impact.tex
\section*{Impact Statement}
This research aims to deepen the understanding of how AT methods influence generalization and robustness, leading to more effective and practical adversarial training techniques. 
As machine learning models become increasingly integrated into real-world applications, it is crucial to develop frameworks that mitigate misuse while ensuring ethical compliance. 
Building on prior work in adversarial training, our approach enhances the robustness of encoder-based models while significantly improving computational efficiency.

%% file: appendix.tex
\newpage
\onecolumn
\section{Supplementary Material}

In this paper, we propose \system{}, an AT algorithm that not only optimizes for better robustness and generalization but also enhances scalability. 
Specifically, \system{} leverages the manifold conjecture, which posits that OFM-AEs lead to better robustness while ONM-AEs enhance generalization. 
To achieve this, \system{} perturbs the layer with the lowest intrinsic dimension (ID). Intuitively, perturbing this layer would yield the highest proportion of OFM-AEs across layers. Formally, \system{} solves the following program:
\begin{equation*}
\min_{\theta}  \mathbb{E}_{(x,y)\sim\mathcal{D}}\Big[ \max_{ \|\delta_{l^*}\| \leq \epsilon_{l^*} }\ell\Big(f_{\theta^{[l^*,n]}}\Big(f_{\theta^{[1,l^*]}}(x)+\delta_{l^*}\Big), y \Big)  \Big].
\label{app:eq:adversarial_loss_approach}
\end{equation*}
We show that $l^*$ corresponds to the last layer in enc-LLMs and the first layer in dec-LLMs. This explains the difference in robustness and generalization trends between vision models and enc-LLMs \cite{Stutz_2019_CVPR,shafahi2019adversarial,Zhu2020FreeLB}. 
Specifically, in vision models, improvements in robustness are often accompanied by a decrease in generalization. In contrast, in enc-LLMs, improvements in generalization are achieved with lesser gains in robustness. The algorithm for \system{} is provided in Alg. \ref{alg:method}.\\

\input{alg}

\noindent\system{} has the advantage of being significantly more runtime-efficient than classical adversarial training. The gain is of the order of $\cO\Big(P(n-l^{*}) \Big(\max\!\Big(\{d_{l_i} | l_i \in [1, l^{*}-1]  \} \Big)-d_{l^*} \Big) \Big)$, for an $n$-layered deepnet under a $P$-step PGD attack. $d_l$ is the dimensionality of the layer $l$.\\

\noindent Empirical results demonstrate that \system{} leads to better robustness while maintaining comparable accuracy to standard training. We achieve compelling results on robustifying (1) sentiment classifiers, (2) safety filters in decoder-based models, and (3) retriever models in the Retrieval Augmented Generation (RAG) setup.\\

\noindent The remaining of the supplementary material is organized as follows:
\begin{itemize}
    \item Appendix \ref{sec:relation}: Intrinsic Dimension Estimation
    \item Appendix \ref{sec:app:lat}: Additional Results of ID \& Robustness/Generalization Relationship
    \item Appendix \ref{sec:app:exp}: Additional Results of Robustifying Sentiment Classifiers
    \item Appendix \ref{sec:app:glue}: Additional Results of Language Understanding Benchmarks
    \item Appendix \ref{sec:app:gcg}: Additional Results of Robustifying Safety Filters In Decoder Based LLMs
\end{itemize}

\input{supplementary}

%% file: alg.tex
\begin{algorithm}[!]
\caption{\system{}}
\begin{algorithmic}[1]
\STATE \textbf{Input:} $\cD=\{\cX, \cY\}$: input data, $f_{\theta}$: a deepnet model, $\epsilon$: attack strength, $E$: the number of epochs, $\alpha$: PGD learning rate, and $\Pi$: the projection operator into the $\epsilon$-ball.
\STATE \textbf{Output} $f_{\theta}$: a deepnet model
\STATE \textit{ \% Determine the ID behaviour of the model with twoNN}
\STATE \textit{ \% Identify the optimal layer $l^{*}$ to perturb (Eq. \eqref{eq:opt_layer_fine}})
\FOR{$e=1,..,E$}
    \FOR{$(x_i,y_i) \in \mathcal{D}$}
        \STATE $\delta_{l^*} \sim \mathcal{N}(0,\sigma^{2}I)$ \textit{ \% sample an initial perturbation} \\
        \STATE \textit{ \% store forward pass for scalability} \\
        \STATE $ mid\_rep \leftarrow  f_{\theta^{[0,l^*]}}(x_i)$ 
        \FOR{$s=1,..,S$} 
            \STATE $loss \leftarrow \ell\left(f_{\theta^{[l^*,n]}}(mid\_rep+\delta_{l^*}),y_i)\right))$ \\
            \STATE $\delta_{l^*}  \leftarrow \Pi_{\epsilon_{l^*}} ( \delta_{l^*} + \alpha \cdot sign\left( \nabla_{\delta_{l^*}}(loss) )\right ) $
        \ENDFOR
        \STATE \textit{ \% update the model parameters} 
        \STATE $loss \leftarrow \ell\left(f_{\theta^{[l^*,n]}}(mid\_rep+\delta_{l^*}),y_i)\right)$
        \STATE $\theta^{[l^*,n]} \leftarrow \theta^{[l^*,n]} - lr \nabla_{\theta^{[l^*,n]}} (loss)$  \\
    \ENDFOR
\ENDFOR
\end{algorithmic}
\label{alg:method}
\end{algorithm}

%% file: supplementary.tex
\section{Intrinsic Dimension Estimation}
\label{sec:relation}

In the following, we provide additional details on the experimental setup on OFM-/ONM- AEs ratio calculation and the ID estimation in \figref{fig:k_effect}.\\
\noindent\textbf{Experimental setup.} We conduct experiments on two enc-LLMs (BERT and RoBERTa), across three different datasets (AGNEWS, IMDB, YELP). Additionally, we evaluated three vision models (ViT~\cite{dosovitskiy2020image}, DeiT~\cite{touvron2021training}, VGG-13~\cite{simonyan2014very}) on the CIFAR-10~\cite{cifar10} dataset and three LLMs (LLaMA-2-7b~\cite{touvron2023llama}, Vicuna~\cite{zheng2024judging}, Mistral-7b~\cite{jiang2023mistral}) on the HelpSteer dataset ~\cite{wang2023helpsteer}.
We utilize the train split of the datasets to estimate the ID and eigenspace of the layers. 
AEs are generated using TextFooler for LMs and PGD for vision models on the respective test splits of the datasets. 
For LLMs, AEs are generated using the Greedy Coordinate Gradient (GCG) attack \cite{zou2023universal} on the harmful prompt datasets released in the same paper.
In our calculation, we use CLS embedding for BERT, RoBERTa, ViT, and DeiT models; last token embedding for LLMs; and conventional layer output for VGG-13.

\noindent\textbf{Discussion of results.} We present the average projection error, $e_l^k$, of each layer on the first row of \figref{fig:k_effect}. 
Results in \figref{fig:k_effect}(a,b) indicate that for LMs, the average $e_l^k$ monotonically increases, suggesting that examples become more off-manifold at the latest layers, consistent with our hypothesis. 
Conversely, for vision models and dec-LLMs in \figref{fig:k_effect}(c,d), we observe the opposite characteristic with a lower average $e_l^k$ at the latest layers as expected.
The only unexpected behavior observed in vision models and dec-LLMs, except for VGG-13 and Mistral, is that they exhibit the highest off-manifold ratio ($e_l^k$) in the initial layers rather than the first layers. For transformer models, the CLS/last token embedding starts with a specific value and gradually evolves to represent the sentence. This means it takes time for the effect of AEs to manifest in the CLS/last token embedding. In contrast, since the VGG model's representation is directly calculated from the input, the highest off-manifold ratio is obtained at the first layer.

We hypothesize that this phenomenon can be explained by the ID of the layers. 
Specifically, if $I_l \ll d_l$, AEs tend to be off-manifold. 
To validate this hypothesis for the same models and dataset, we measure $I_l$ using the twoNN method and normalize it with $d_l$. The normalized $I_l$ can be seen in the second row of \figref{fig:k_effect}.
In line with our hypothesis, while it decreases for the LMs (\figref{fig:k_effect}(a,b)), it increases for vision models and dec-LLMs(\figref{fig:k_effect}(c,d)).

\section{Additional Results of ID \& Robustness/Generalization Relationship }
\label{sec:app:lat}

We further examine the relationship between intrinsic dimensionality (ID), robustness, and generalization by applying Latent Adversarial Training (LAT) to different layers of the LLaMA-2-7B model. 
Unlike untargeted LAT, which disrupts model behavior, we employ targeted LAT to induce specific adversarial responses, following \citet{casper2024defending}.
This is achieved by perturbing the residual stream with L2-norm-bounded noise using PGD.  

To stabilize training and mitigate unintended effects, we interleave LAT with supervised fine-tuning on the UltraChat dataset \cite{ding2023enhancing}. We evaluate generalization using MT-Bench \cite{zheng2023judging} and robustness using GCG-based and PAIR attacks from HarmBench \cite{mazeika2024harmbench}. 
The model is trained with a learning rate of $2e-4$, applying LAT at every even-numbered layer with norm bounds ranging from 1 to 5.  

The results are presented in Figure~\ref{fig:llama-attack}. 
Attacking lower layers (light blue markers) improves robustness but reduces generalization, as indicated by points concentrated in the bottom-right region. 
In contrast, attacking upper layers (dark blue markers) enhances generalization at the cost of robustness, shifting results toward the top-left. Additionally, while model robustness drops to as low as 30\% under the GCG attack, it remains consistently above 75\% across all configurations under the PAIR attack.
\section{Additional Results of Robustifying Sentiment Classifiers}
\label{sec:app:exp}

 \input{figures/llama-attack}

In the following, we provide more details about the experiments on robustifying sentiment classifiers.\\
\textbf{Datasets.} 
We evaluate \system on three datasets: AG-News Corpus (AGNEWS) \cite{zhang2015character}, Internet Movie Database (IMDB) \cite{maas-etal-2011-learning}, and Yelp Review Polarity (YELP) \cite{zhang2015character}. The AGNEWS dataset contains over 120000 samples, each belonging to one of the four labels: World, Sports, Business, Sci/Tech. The IMDB dataset contains 50000 data samples of movie reviews with binary labels for negative and positive sentiments. The YELP dataset contains nearly 600000 samples of highly polar Yelp reviews with binary labels. However, due to limitations in computing resources, we only use a subset of 63000 samples of the YELP dataset. In addition, we randomly sample 10\% of the training set for validation in all datasets. For testing, we use a subset of 1000 test samples from each dataset, following previous work practices.
The AGNEWS dataset contains over 120k samples, categorized into four classes: World, Sports, Business, and Sci/Tech. 
The IMDB dataset consists of 50k movie reviews, each labeled with binary sentiments (positive or negative). \\
\noindent\textbf{Base model.}
We employed the $\text{BERT}_{\text{base-cased}}$ \cite{devlin-etal-2019-bert} and $\text{RoBERTa}_{\text{base-cased}}$ \cite{liu2019roberta} models in our experiments. 
To conduct the evaluations, we utilize the fine-tuned models provided by \emph{TextAttack} from HuggingFace for all datasets, except for the RoBERTa base model fine-tuned on YELP dataset. 
For the YELP dataset, we created a fine-tuned RoBERTa model for 2 epochs with a learning rate of $1e-05$ and a batch size of 32. \\
\textbf{Adversarial Attacks.}
which include the following constraints: (1) The maximum percentage of modified words is set to $0.3$ for AGNEWS, $0.1$ for IMDB and YELP datasets, respectively. 
(2) For word replacement, a maximum of $50$ candidates are considered for each word. (3) The semantic similarity, measured using the Universal Sentence Encoder \cite{cer-etal-2018-universal}, between the original input and the generated adversarial example must exceed $0.84$. PWWS uses word synonyms, TextFooler applies nearest neighbor search in counter-fitting embeddings \cite{mrksic-etal-2016-counter}, and BERT-Attack utilizes BERT masked language model to generate candidate words.\\
\textbf{Baselines.}
For input space adversarial training, we consider Adversarial Sparse Convex Combination (ASCC) \cite{DBLP:conf/iclr/DongLJ021} which model the perturbation space as the convex hull of word synonyms. 
ASCC incorporates an entropy-based sparsity regularizer to capture word substitution geometry more effectively.
In our investigation of embedding space adversarial training which recognized as the most impactful technique for enhancing generalization~\cite{li2021searching}, we conduct a thorough analysis of FreeLB++~\cite{li-etal-2021-searching} (employs gradient-guided perturbations centered around the most susceptible data points).
For certified defenses, we evaluate SAFER \cite{ye-etal-2020-safer}, TMD \cite{minh2022textual}, and RSMI \cite{minh2022textual}. 
SAFER constructs a set of randomized inputs by performing random synonym substitutions and using the statistical properties of predicted labels to certify robustness. 
TMD employs infoGAN \cite{DBLP:conf/nips/ChenCDHSSA16} to project adversarial examples to the data manifold in the last layer to address the manifold issue. 
RMSI combines these ideas by applying importance-based masking to tokens and leveraging randomized smoothing in each layer.

\noindent\textbf{Implementation details.}
To train the last layer of $f_{\theta}$ with adversarial samples, we create adversarial samples using 5-step PGD attacks. During training, we use epsilon values of $0.1$, $0.1$, and $0.8$ for the YELP, AGNEWS, and IMDB datasets, respectively, for the BERT models. For the RoBERTa models, we employ epsilon values of $0.1$, $0.6$, and $0.03$. All models are trained $10$ epochs with a learning rate of $0.1$. In our evaluation, we use a V100 GPU with 32 GB memory and 64 CPUs. %

\section{Additional Results of Language Understanding Benchmarks}
\label{sec:app:glue}

\input{Tables/tab_glue}

\input{Tables/gcg_app}

To comprehensively evaluate \system{}'s performance against a broader spectrum of textual adversarial attacks, we employ the GLUE and AdvGLUE benchmarks.
The GLUE benchmark \cite{wang2019glue} is a comprehensive evaluation suite featuring seven diverse NLP tasks to assess model performance. 
The AdvGLUE benchmark \cite{wang2021adversarial} is an extension of GLUE, incorporating 17 distinct textual adversarial attacks, covering word-level transformations, sentence-level manipulations, and human-written AEs. 
This extension ensures a thorough evaluation encompassing various adversarial linguistic phenomena.
For our assessment, we employ the evaluation sets of four datasets across three different tasks: Sentiment Analysis (SST-2), Duplicate Question Detection (QQP), and Natural Language Inference (QNLI, RTE).

In our evaluation, we compare SMAAT against standard BERT and RoBERTa models\footnote{We use the fine-tuned models available from \textit{https://huggingface.co/JeremiahZ}}, as well as their FreeLB++ incorporated versions.
In the case of \system{}, we conducted a grid search for the learning rate, ranging from $0.1$ to $0.001$, and the $\epsilon$ value, ranging from $0.8$ to $0.01$, using 3-PGD steps
As shown in Table \ref{tab:glue}, SMAAT demonstrates a robustness improvement of 5.6\% and 2.6\% for BERT, and 12.1\% and 2.5\% for RoBERTa, compared to the standard and FreeLB++ models, respectively, while maintaining similar generalization performance.

\section{Additional Results of Robustifying Safety Filters In Decoder Based LLMs}
\label{sec:app:gcg}

The GCG attack is deliberately designed to bypass the safety alignment of LLMs by generating a response to a potentially harmful prompt through appending an adversarial suffix to the user's input. This strategy leverages previous methodologies by using (1) an affirmative response tactic \cite{wei2023jailbroken,carlini2023are} to direct the model's output towards the attacker's intended outcome (used for loss calculation) and (2) a mix of greedy and gradient-based discrete optimization \cite{shin2020autoprompt} to pinpoint the most susceptible tokens. A key attribute of the GCG attack is its transferability, demonstrating that adversarial suffix designed for a specific prompt on one model can successfully affect a broad range of other models.

A suggested defense against the GCG attack involves incorporating a lightweight binary classifier model designed to identify harmful prompts \cite{kumar2023certifying}. 
It is important to note, however, that these classifiers can still be vulnerable to such attacks.
To assess the effectiveness of \system{} against the GCG attack, we conduct the attack by setting the suffix length to 20 tokens.
The attack is generated over 50 iterations, with 100 trials per iteration. 
Rather than leveraging the attack's transferability, we tailor an adversarial suffix for every individual prompt and model to enhance the attack's impact.
Our assessment is conducted using two datasets, AdvBench \cite{zou2023universal} and HH-RLHF \cite{bai2022training}, and involves three models: BERT, RoBERTa, and DistilBERT. AdvBench features 640 training prompts (320 harmful, 320 safe) and 240 test prompts (120 harmful, 120 safe), whereas HH-RLHF includes approximately 44K harmful and 44K helpful training prompts, with 2.3K test samples for each prompt category.
In all cases, standard models are trained over 5 epochs with a learning rate of $1e^{-5}$. 
Table~\ref{tab:gcg_hyper} details the training hyperparameters for \system{}.

%% file: figures/llama-attack.tex
\begin{figure*}
  \begin{center}
    \begin{minipage}[a]{0.48\textwidth}
    \centering
    \includegraphics[width=0.8\columnwidth]{figures/figures_files/llama2_gen_robust.pdf}
    \centerline{(a) GCG }
  \end{minipage}
  \begin{minipage}[a]{0.48\textwidth}
    \centering
    \includegraphics[width=0.8\columnwidth]{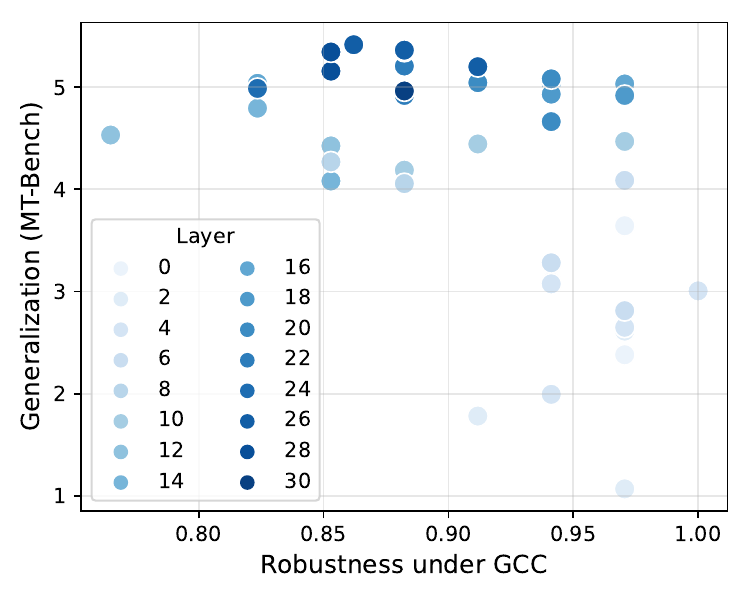}
    \centerline{(b) PAIR}
  \end{minipage}
\caption{Layer-wise effects of adversarial training on generalization and robustness against GCG and PAIR attacks on the LLaMA-2 model.
Marker colors represent layer depth, transitioning from light blue (lower layers) to dark blue (higher layers). The results show that robustness drops to as low as 30\% under the GCG attack, while remaining above 75\% under the PAIR attack across all configurations.}
  \label{fig:llama-attack}
  \end{center}
\end{figure*}

%% file: Tables/tab_glue.tex
\begin{table*}
\caption{Average accuracy from the standard training, FreeLB++ and SMAAT on GLUE and AdvGLUE datasets. 
Results clearly demonstrate that \system{} enhances model generalization (GLUE results) and robustness (AdvGLUE results).} 
\centering
\begin{tabular}{c|ccc|ccc}
\toprule 
\multirow{2}{*}{Dataset}   & \multicolumn{3}{c|}{BERT}               & \multicolumn{3}{c}{RoBERTa}       \\ %
                         & Standard & FreeLB++ & \system{} (Ours)  & Standard & FreeLB++ & \system{} (Ours)         \\ 
    \midrule
 GLUE        & 85.9 & 86.3 & 86.3  & 89.3  & 89.6  & 89.7     \\ 
 AdvGLUE     & 39.5 & 42.5 & 45.1  & 27.5  & 37.1  & 39.6      \\

    \bottomrule
                                                                                             
\end{tabular}%

\label{tab:glue}
             
\end{table*}

%% file: Tables/gcg_app.tex
\begin{table*}
  \centering
    \caption{Hyperparameters in experiments on robustifying the safety filters on decoding LLMs experiments. Learning rate (lr) and $\epsilon$ values for \system{} and FreeLB++ methods.}
    \begin{tabular}{c|cc|cc|cc||cc|cc|cc}
    \toprule 
     \multirow{3}{*}{Dataset}   & \multicolumn{6}{c||}{FreeLB++}& \multicolumn{6}{c}{\system{}} \\
     & \multicolumn{2}{c|}{BERT}               & \multicolumn{2}{c}{RoBERTa}    & \multicolumn{2}{c||}{DistilBERT}  & \multicolumn{2}{c|}{BERT}               & \multicolumn{2}{c}{RoBERTa}    & \multicolumn{2}{c}{DistilBERT}       \\ %
                           & lr & $\epsilon$ & $lr$ & $\epsilon$ &  $lr$ & $\epsilon$  & lr & $\epsilon$ & $lr$ & $\epsilon$ &  $lr$ & $\epsilon$           \\ 
      \midrule
       AdvBench            & 5e-3 & 0.05  & 5e-3 & 0.05 & 1e-2  & 0.01    &1e-4&1e-3&1e-4&1e-3&1e-4&1e-4    \\ 
       HH-RLHF            & 5e-3  & 0.5  & 53-3  & 0.05  & 5e-4 & 	0.05  &1e-4&5e-4&1e-4&1e-3&1e-4&5e-4     \\ 
     \bottomrule                                     
    \end{tabular}%
    \label{tab:gcg_hyper}
\end{table*}

%% file: main.bbl
\begin{thebibliography}{79}
\providecommand{\natexlab}[1]{#1}
\providecommand{\url}[1]{\texttt{#1}}
\expandafter\ifx\csname urlstyle\endcsname\relax
  \providecommand{\doi}[1]{doi: #1}\else
  \providecommand{\doi}{doi: \begingroup \urlstyle{rm}\Url}\fi

\bibitem[Alemany \& Pissinou(2020)Alemany and Pissinou]{alemany2020dilemma}
Alemany, S. and Pissinou, N.
\newblock The dilemma between data transformations and adversarial robustness for time series application systems.
\newblock In \emph{AAAI}, 2020.

\bibitem[Altinisik et~al.(2023{\natexlab{a}})Altinisik, Messaoud, Sencar, and Chawla]{altinisik2023a3t}
Altinisik, E., Messaoud, S., Sencar, H.~T., and Chawla, S.
\newblock A3t: accuracy aware adversarial training.
\newblock \emph{Machine Learning}, pp.\  1--20, 2023{\natexlab{a}}.

\bibitem[Altinisik et~al.(2023{\natexlab{b}})Altinisik, Sajjad, Sencar, Messaoud, and Chawla]{altinisik2022impact}
Altinisik, E., Sajjad, H., Sencar, H., Messaoud, S., and Chawla, S.
\newblock Impact of adversarial training on robustness and generalizability of language models.
\newblock In \emph{Findings of the ACL}, 2023{\natexlab{b}}.

\bibitem[Bai et~al.(2021)Bai, Luo, Zhao, Wen, and Wang]{ijcai2021p591}
Bai, T., Luo, J., Zhao, J., Wen, B., and Wang, Q.
\newblock Recent advances in adversarial training for adversarial robustness.
\newblock In \emph{IJCAI-21}, 2021.

\bibitem[Bai et~al.(2022)Bai, Jones, Ndousse, Askell, Chen, DasSarma, Drain, Fort, Ganguli, Henighan, et~al.]{bai2022training}
Bai, Y., Jones, A., Ndousse, K., Askell, A., Chen, A., DasSarma, N., Drain, D., Fort, S., Ganguli, D., Henighan, T., et~al.
\newblock Training a helpful and harmless assistant with reinforcement learning from human feedback.
\newblock \emph{arXiv preprint arXiv:2204.05862}, 2022.

\bibitem[Cao et~al.(2023)Cao, Cao, Lin, and Chen]{cao2023defending}
Cao, B., Cao, Y., Lin, L., and Chen, J.
\newblock Defending against alignment-breaking attacks via robustly aligned llm, 2023.

\bibitem[Carlini et~al.(2023)Carlini, Nasr, Choquette-Choo, Jagielski, Gao, Koh, Ippolito, Tram{\`e}r, and Schmidt]{carlini2023are}
Carlini, N., Nasr, M., Choquette-Choo, C.~A., Jagielski, M., Gao, I., Koh, P.~W., Ippolito, D., Tram{\`e}r, F., and Schmidt, L.
\newblock Are aligned neural networks adversarially aligned?
\newblock In \emph{NeurIPS}, 2023.

\bibitem[Casper et~al.(2024)Casper, Schulze, Patel, and Hadfield-Menell]{casper2024defending}
Casper, S., Schulze, L., Patel, O., and Hadfield-Menell, D.
\newblock Defending against unforeseen failure modes with latent adversarial training.
\newblock \emph{arXiv preprint arXiv:2403.05030}, 2024.

\bibitem[Cer et~al.(2018)Cer, Yang, Kong, Hua, Limtiaco, St.~John, Constant, Guajardo-Cespedes, Yuan, Tar, Strope, and Kurzweil]{cer-etal-2018-universal}
Cer, D., Yang, Y., Kong, S.-y., Hua, N., Limtiaco, N., St.~John, R., Constant, N., Guajardo-Cespedes, M., Yuan, S., Tar, C., Strope, B., and Kurzweil, R.
\newblock Universal sentence encoder for {E}nglish.
\newblock In \emph{EMNLP}, 2018.

\bibitem[Chao et~al.(2023)Chao, Robey, Dobriban, Hassani, Pappas, and Wong]{chao2023jailbreaking}
Chao, P., Robey, A., Dobriban, E., Hassani, H., Pappas, G.~J., and Wong, E.
\newblock Jailbreaking black box large language models in twenty queries.
\newblock \emph{arXiv preprint arXiv:2310.08419}, 2023.

\bibitem[Chen et~al.(2016)Chen, Duan, Houthooft, Schulman, Sutskever, and Abbeel]{DBLP:conf/nips/ChenCDHSSA16}
Chen, X., Duan, Y., Houthooft, R., Schulman, J., Sutskever, I., and Abbeel, P.
\newblock Infogan: Interpretable representation learning by information maximizing generative adversarial nets.
\newblock In \emph{NeurIPS}, 2016.

\bibitem[Cheng et~al.(2022)Cheng, Lei, Chen, Dhillon, and Hsieh]{cheng2020cat}
Cheng, M., Lei, Q., Chen, P., Dhillon, I.~S., and Hsieh, C.
\newblock {CAT:} customized adversarial training for improved robustness.
\newblock In \emph{IJCAI}, pp.\  673--679, 2022.

\bibitem[Devlin et~al.(2019)Devlin, Chang, Lee, and Toutanova]{devlin-etal-2019-bert}
Devlin, J., Chang, M.-W., Lee, K., and Toutanova, K.
\newblock {BERT}: Pre-training of deep bidirectional transformers for language understanding.
\newblock In \emph{ACL}, 2019.

\bibitem[Ding et~al.(2023)Ding, Chen, Xu, Qin, Zheng, Hu, Liu, Sun, and Zhou]{ding2023enhancing}
Ding, N., Chen, Y., Xu, B., Qin, Y., Zheng, Z., Hu, S., Liu, Z., Sun, M., and Zhou, B.
\newblock Enhancing chat language models by scaling high-quality instructional conversations, 2023.

\bibitem[Dong et~al.(2021)Dong, Luu, Ji, and Liu]{DBLP:conf/iclr/DongLJ021}
Dong, X., Luu, A.~T., Ji, R., and Liu, H.
\newblock Towards robustness against natural language word substitutions.
\newblock In \emph{ICLR}, 2021.

\bibitem[Dosovitskiy et~al.(2020)Dosovitskiy, Beyer, Kolesnikov, Weissenborn, Zhai, Unterthiner, Dehghani, Minderer, Heigold, Gelly, et~al.]{dosovitskiy2020image}
Dosovitskiy, A., Beyer, L., Kolesnikov, A., Weissenborn, D., Zhai, X., Unterthiner, T., Dehghani, M., Minderer, M., Heigold, G., Gelly, S., et~al.
\newblock An image is worth 16x16 words: Transformers for image recognition at scale.
\newblock \emph{arXiv preprint arXiv:2010.11929}, 2020.

\bibitem[Ethayarajh(2019)]{ethayarajh-2019-contextual}
Ethayarajh, K.
\newblock How contextual are contextualized word representations? {C}omparing the geometry of {BERT}, {ELM}o, and {GPT}-2 embeddings.
\newblock In \emph{IJCNLP}, 2019.

\bibitem[Facco et~al.(2017)Facco, d’Errico, Rodriguez, and Laio]{facco2017estimating}
Facco, E., d’Errico, M., Rodriguez, A., and Laio, A.
\newblock Estimating the intrinsic dimension of datasets by a minimal neighborhood information.
\newblock \emph{Nature Scientific reports}, 2017.

\bibitem[Fatehkia et~al.(2025)Fatehkia, Altinisik, and Sencar]{SGM}
Fatehkia, M., Altinisik, E., and Sencar, H.~T.
\newblock Sgm: A framework for building specification-guided moderation filters.
\newblock \emph{arXiv preprint}, 2025.

\bibitem[Gan et~al.(2020)Gan, Chen, Li, Zhu, Cheng, and Liu]{villa}
Gan, Z., Chen, Y., Li, L., Zhu, C., Cheng, Y., and Liu, J.
\newblock Large-scale adversarial training for vision-and-language representation learning.
\newblock In \emph{NeurIPS}, 2020.

\bibitem[Gilmer et~al.(2018)Gilmer, Metz, Faghri, Schoenholz, Raghu, Wattenberg, and Goodfellow]{DBLP:conf/iclr/GilmerMFSRWG18}
Gilmer, J., Metz, L., Faghri, F., Schoenholz, S.~S., Raghu, M., Wattenberg, M., and Goodfellow, I.~J.
\newblock Adversarial spheres.
\newblock In \emph{ICLR}, 2018.

\bibitem[Izacard et~al.(2021)Izacard, Caron, Hosseini, Riedel, Bojanowski, Joulin, and Grave]{izacard2021unsupervised}
Izacard, G., Caron, M., Hosseini, L., Riedel, S., Bojanowski, P., Joulin, A., and Grave, E.
\newblock Unsupervised dense information retrieval with contrastive learning.
\newblock \emph{arXiv preprint arXiv:2112.09118}, 2021.

\bibitem[Jiang et~al.(2023)Jiang, Sablayrolles, Mensch, Bamford, Chaplot, Casas, Bressand, Lengyel, Lample, Saulnier, et~al.]{jiang2023mistral}
Jiang, A.~Q., Sablayrolles, A., Mensch, A., Bamford, C., Chaplot, D.~S., Casas, D. d.~l., Bressand, F., Lengyel, G., Lample, G., Saulnier, L., et~al.
\newblock Mistral 7b.
\newblock \emph{arXiv preprint arXiv:2310.06825}, 2023.

\bibitem[Jin et~al.(2020{\natexlab{a}})Jin, Jin, Zhou, and Szolovits]{Jin_Jin_Zhou_Szolovits_2020}
Jin, D., Jin, Z., Zhou, J.~T., and Szolovits, P.
\newblock Is bert really robust? a strong baseline for natural language attack on text classification and entailment.
\newblock In \emph{AAAI}, 2020{\natexlab{a}}.

\bibitem[Jin et~al.(2020{\natexlab{b}})Jin, Jin, Zhou, and Szolovits]{jin2020bert}
Jin, D., Jin, Z., Zhou, J.~T., and Szolovits, P.
\newblock Is bert really robust? a strong baseline for natural language attack on text classification and entailment.
\newblock In \emph{AAAI}, 2020{\natexlab{b}}.

\bibitem[Krizhevsky(2009)]{cifar10}
Krizhevsky, A.
\newblock Learning multiple layers of features from tiny images.
\newblock Technical report, 2009.

\bibitem[Kumar et~al.(2023)Kumar, Agarwal, Srinivas, Li, Feizi, and Lakkaraju]{kumar2023certifying}
Kumar, A., Agarwal, C., Srinivas, S., Li, A.~J., Feizi, S., and Lakkaraju, H.
\newblock Certifying llm safety against adversarial prompting, 2023.

\bibitem[Kurakin et~al.(2017)Kurakin, Goodfellow, and Bengio]{kurakin2018adversarial}
Kurakin, A., Goodfellow, I.~J., and Bengio, S.
\newblock Adversarial examples in the physical world.
\newblock In \emph{ICLR}, 2017.

\bibitem[Kwiatkowski et~al.(2019)Kwiatkowski, Palomaki, Redfield, Collins, Parikh, Alberti, Epstein, Polosukhin, Devlin, Lee, et~al.]{kwiatkowski2019natural}
Kwiatkowski, T., Palomaki, J., Redfield, O., Collins, M., Parikh, A., Alberti, C., Epstein, D., Polosukhin, I., Devlin, J., Lee, K., et~al.
\newblock Natural questions: a benchmark for question answering research.
\newblock In \emph{ACL}, 2019.

\bibitem[Li et~al.(2020{\natexlab{a}})Li, Ma, Guo, Xue, and Qiu]{li-etal-2020-bert-attack}
Li, L., Ma, R., Guo, Q., Xue, X., and Qiu, X.
\newblock {BERT}-{ATTACK}: Adversarial attack against {BERT} using {BERT}.
\newblock In \emph{EMNLP}, 2020{\natexlab{a}}.

\bibitem[Li et~al.(2020{\natexlab{b}})Li, Ma, Guo, Xue, and Qiu]{li2020bert}
Li, L., Ma, R., Guo, Q., Xue, X., and Qiu, X.
\newblock Bert-attack: Adversarial attack against bert using bert.
\newblock \emph{arXiv preprint arXiv:2004.09984}, 2020{\natexlab{b}}.

\bibitem[Li et~al.(2021{\natexlab{a}})Li, Min, Lee, Yu, Kruus, Wang, and Hsieh]{Li_2021_ICCV}
Li, Y., Min, M.~R., Lee, T., Yu, W., Kruus, E., Wang, W., and Hsieh, C.-J.
\newblock Towards robustness of deep neural networks via regularization.
\newblock In \emph{ICCV}, 2021{\natexlab{a}}.

\bibitem[Li et~al.(2021{\natexlab{b}})Li, Xu, Zeng, Li, Zheng, Zhang, Chang, and Hsieh]{li-etal-2021-searching}
Li, Z., Xu, J., Zeng, J., Li, L., Zheng, X., Zhang, Q., Chang, K.-W., and Hsieh, C.-J.
\newblock Searching for an effective defender: Benchmarking defense against adversarial word substitution.
\newblock In \emph{EMNLP}, 2021{\natexlab{b}}.

\bibitem[Li et~al.(2021{\natexlab{c}})Li, Xu, Zeng, Li, Zheng, Zhang, Chang, and Hsieh]{li2021searching}
Li, Z., Xu, J., Zeng, J., Li, L., Zheng, X., Zhang, Q., Chang, K.-W., and Hsieh, C.-J.
\newblock Searching for an effective defender: Benchmarking defense against adversarial word substitution.
\newblock \emph{arXiv preprint arXiv:2108.12777}, 2021{\natexlab{c}}.

\bibitem[Liu et~al.(2023)Liu, Xu, Chen, and Xiao]{liu2023autodan}
Liu, X., Xu, N., Chen, M., and Xiao, C.
\newblock Autodan: Generating stealthy jailbreak prompts on aligned large language models.
\newblock \emph{arXiv preprint arXiv:2310.04451}, 2023.

\bibitem[Liu et~al.(2019)Liu, Ott, Goyal, Du, Joshi, Chen, Levy, Lewis, Zettlemoyer, and Stoyanov]{liu2019roberta}
Liu, Y., Ott, M., Goyal, N., Du, J., Joshi, M., Chen, D., Levy, O., Lewis, M., Zettlemoyer, L., and Stoyanov, V.
\newblock Roberta: A robustly optimized bert pretraining approach.
\newblock \emph{arXiv preprint arXiv:1907.11692}, 2019.

\bibitem[Ma et~al.(2018)Ma, Li, Wang, Erfani, Wijewickrema, Schoenebeck, Houle, Song, and Bailey]{ma2018characterizing}
Ma, X., Li, B., Wang, Y., Erfani, S.~M., Wijewickrema, S., Schoenebeck, G., Houle, M.~E., Song, D., and Bailey, J.
\newblock Characterizing adversarial subspaces using local intrinsic dimensionality.
\newblock In \emph{ICLR}, 2018.

\bibitem[Maas et~al.(2011)Maas, Daly, Pham, Huang, Ng, and Potts]{maas-etal-2011-learning}
Maas, A.~L., Daly, R.~E., Pham, P.~T., Huang, D., Ng, A.~Y., and Potts, C.
\newblock Learning word vectors for sentiment analysis.
\newblock In \emph{ACL}, 2011.

\bibitem[Madry et~al.(2017)Madry, Makelov, Schmidt, Tsipras, and Vladu]{madry2017towards}
Madry, A., Makelov, A., Schmidt, L., Tsipras, D., and Vladu, A.
\newblock Towards deep learning models resistant to adversarial attacks.
\newblock \emph{arXiv preprint arXiv:1706.06083}, 2017.

\bibitem[Mazeika et~al.(2024)Mazeika, Phan, Yin, Zou, Wang, Mu, Sakhaee, Li, Basart, Li, Forsyth, and Hendrycks]{mazeika2024harmbench}
Mazeika, M., Phan, L., Yin, X., Zou, A., Wang, Z., Mu, N., Sakhaee, E., Li, N., Basart, S., Li, B., Forsyth, D., and Hendrycks, D.
\newblock Harmbench: A standardized evaluation framework for automated red teaming and robust refusal.
\newblock 2024.

\bibitem[Meng \& Chen(2017)Meng and Chen]{meng2017magnet}
Meng, D. and Chen, H.
\newblock Magnet: a two-pronged defense against adversarial examples.
\newblock In \emph{ACM CCS}, 2017.

\bibitem[Minh \& Luu(2022)Minh and Luu]{minh2022textual}
Minh, D.~N. and Luu, A.~T.
\newblock Textual manifold-based defense against natural language adversarial examples.
\newblock In \emph{EMNLP}, 2022.

\bibitem[Morris et~al.(2020)Morris, Lifland, Yoo, Grigsby, Jin, and Qi]{morris-etal-2020-textattack}
Morris, J., Lifland, E., Yoo, J.~Y., Grigsby, J., Jin, D., and Qi, Y.
\newblock {T}ext{A}ttack: A framework for adversarial attacks, data augmentation, and adversarial training in {NLP}.
\newblock In \emph{EMNLP}, 2020.

\bibitem[Mrk{\v{s}}i{\'c} et~al.(2016)Mrk{\v{s}}i{\'c}, {\'O}~S{\'e}aghdha, Thomson, Ga{\v{s}}i{\'c}, Rojas-Barahona, Su, Vandyke, Wen, and Young]{mrksic-etal-2016-counter}
Mrk{\v{s}}i{\'c}, N., {\'O}~S{\'e}aghdha, D., Thomson, B., Ga{\v{s}}i{\'c}, M., Rojas-Barahona, L.~M., Su, P.-H., Vandyke, D., Wen, T.-H., and Young, S.
\newblock Counter-fitting word vectors to linguistic constraints.
\newblock In \emph{ACL}, 2016.

\bibitem[Ren et~al.(2019)Ren, Deng, He, and Che]{ren-etal-2019-generating}
Ren, S., Deng, Y., He, K., and Che, W.
\newblock Generating natural language adversarial examples through probability weighted word saliency.
\newblock In \emph{ACL}, 2019.

\bibitem[Samangouei et~al.(2018)Samangouei, Kabkab, and Chellappa]{samangouei2018defensegan}
Samangouei, P., Kabkab, M., and Chellappa, R.
\newblock Defense-{GAN}: Protecting classifiers against adversarial attacks using generative models.
\newblock In \emph{ICLR}, 2018.

\bibitem[Schott et~al.(2018)Schott, Rauber, Bethge, and Brendel]{schott2018towards}
Schott, L., Rauber, J., Bethge, M., and Brendel, W.
\newblock Towards the first adversarially robust neural network model on mnist.
\newblock \emph{arXiv preprint arXiv:1805.09190}, 2018.

\bibitem[Shafahi et~al.(2019)Shafahi, Najibi, Ghiasi, Xu, Dickerson, Studer, Davis, Taylor, and Goldstein]{shafahi2019adversarial}
Shafahi, A., Najibi, M., Ghiasi, M.~A., Xu, Z., Dickerson, J., Studer, C., Davis, L.~S., Taylor, G., and Goldstein, T.
\newblock Adversarial training for free!
\newblock \emph{Advances in Neural Information Processing Systems}, 32, 2019.

\bibitem[Shamir et~al.(2021)Shamir, Melamed, and BenShmuel]{shamir2021dimpled}
Shamir, A., Melamed, O., and BenShmuel, O.
\newblock The dimpled manifold model of adversarial examples in machine learning.
\newblock \emph{arXiv preprint arXiv:2106.10151}, 2021.

\bibitem[Sheshadri et~al.(2024)Sheshadri, Ewart, Guo, Lynch, Wu, Hebbar, Sleight, Stickland, Perez, Hadfield-Menell, et~al.]{sheshadri2024latent}
Sheshadri, A., Ewart, A., Guo, P., Lynch, A., Wu, C., Hebbar, V., Sleight, H., Stickland, A.~C., Perez, E., Hadfield-Menell, D., et~al.
\newblock Latent adversarial training improves robustness to persistent harmful behaviors in llms.
\newblock \emph{arXiv preprint arXiv:2407.15549}, 2024.

\bibitem[Shin et~al.(2020)Shin, Razeghi, Logan~IV, Wallace, and Singh]{shin2020autoprompt}
Shin, T., Razeghi, Y., Logan~IV, R.~L., Wallace, E., and Singh, S.
\newblock Autoprompt: Eliciting knowledge from language models with automatically generated prompts.
\newblock \emph{arXiv preprint arXiv:2010.15980}, 2020.

\bibitem[Simonyan \& Zisserman(2014)Simonyan and Zisserman]{simonyan2014very}
Simonyan, K. and Zisserman, A.
\newblock Very deep convolutional networks for large-scale image recognition.
\newblock \emph{arXiv preprint arXiv:1409.1556}, 2014.

\bibitem[Song et~al.(2017)Song, Kim, Nowozin, Ermon, and Kushman]{song2017pixeldefend}
Song, Y., Kim, T., Nowozin, S., Ermon, S., and Kushman, N.
\newblock Pixeldefend: Leveraging generative models to understand and defend against adversarial examples.
\newblock \emph{arXiv preprint arXiv:1710.10766}, 2017.

\bibitem[Stewart(1993)]{stewart1993early}
Stewart, G.~W.
\newblock On the early history of the singular value decomposition.
\newblock \emph{SIAM review}, 1993.

\bibitem[Stutz et~al.(2019{\natexlab{a}})Stutz, Hein, and Schiele]{Stutz_2019_CVPR}
Stutz, D., Hein, M., and Schiele, B.
\newblock Disentangling adversarial robustness and generalization.
\newblock In \emph{CVPR}, 2019{\natexlab{a}}.

\bibitem[Stutz et~al.(2019{\natexlab{b}})Stutz, Hein, and Schiele]{stutz2019disentangling}
Stutz, D., Hein, M., and Schiele, B.
\newblock Disentangling adversarial robustness and generalization.
\newblock In \emph{ICCV}, pp.\  6976--6987, 2019{\natexlab{b}}.

\bibitem[Tanay \& Griffin(2016)Tanay and Griffin]{tanay2016boundary}
Tanay, T. and Griffin, L.
\newblock A boundary tilting persepective on the phenomenon of adversarial examples.
\newblock \emph{arXiv preprint arXiv:1608.07690}, 2016.

\bibitem[Touvron et~al.(2021)Touvron, Cord, Douze, Massa, Sablayrolles, and J{\'e}gou]{touvron2021training}
Touvron, H., Cord, M., Douze, M., Massa, F., Sablayrolles, A., and J{\'e}gou, H.
\newblock Training data-efficient image transformers \& distillation through attention.
\newblock In \emph{ICML}, pp.\  10347--10357, 2021.

\bibitem[Touvron et~al.(2023)Touvron, Martin, Stone, Albert, Almahairi, Babaei, Bashlykov, Batra, Bhargava, Bhosale, et~al.]{touvron2023llama}
Touvron, H., Martin, L., Stone, K., Albert, P., Almahairi, A., Babaei, Y., Bashlykov, N., Batra, S., Bhargava, P., Bhosale, S., et~al.
\newblock Llama 2: Open foundation and fine-tuned chat models.
\newblock \emph{arXiv preprint arXiv:2307.09288}, 2023.

\bibitem[Tram{\`{e}}r et~al.(2018)Tram{\`{e}}r, Kurakin, Papernot, Goodfellow, et~al.]{tramer2017ensemble}
Tram{\`{e}}r, F., Kurakin, A., Papernot, N., Goodfellow, I.~J., et~al.
\newblock Ensemble adversarial training: Attacks and defenses.
\newblock In \emph{ICLR}, 2018.

\bibitem[Tsipras et~al.(2019)Tsipras, Santurkar, Engstrom, Turner, and Madry]{tsipras2018robustness}
Tsipras, D., Santurkar, S., Engstrom, L., Turner, A., and Madry, A.
\newblock Robustness may be at odds with accuracy.
\newblock In \emph{ICLR}, 2019.

\bibitem[Wang et~al.(2019{\natexlab{a}})Wang, Singh, Michael, Hill, Levy, and Bowman]{wang2019glue}
Wang, A., Singh, A., Michael, J., Hill, F., Levy, O., and Bowman, S.~R.
\newblock Glue: A multi-task benchmark and analysis platform for natural language understanding.
\newblock In \emph{ICLR}, 2019{\natexlab{a}}.

\bibitem[Wang et~al.(2021)Wang, Xu, Wang, Gan, Cheng, Gao, Awadallah, and Li]{wang2021adversarial}
Wang, B., Xu, C., Wang, S., Gan, Z., Cheng, Y., Gao, J., Awadallah, A.~H., and Li, B.
\newblock Adversarial {GLUE}: A multi-task benchmark for robustness evaluation of language models.
\newblock In \emph{NeurIPS}, 2021.

\bibitem[Wang et~al.(2019{\natexlab{b}})Wang, Zou, Yi, Bailey, Ma, and Gu]{wang2019improving}
Wang, Y., Zou, D., Yi, J., Bailey, J., Ma, X., and Gu, Q.
\newblock Improving adversarial robustness requires revisiting misclassified examples.
\newblock In \emph{ICLR}, 2019{\natexlab{b}}.

\bibitem[Wang et~al.(2023)Wang, Dong, Zeng, Adams, Sreedhar, Egert, Delalleau, Scowcroft, Kant, Swope, et~al.]{wang2023helpsteer}
Wang, Z., Dong, Y., Zeng, J., Adams, V., Sreedhar, M.~N., Egert, D., Delalleau, O., Scowcroft, J.~P., Kant, N., Swope, A., et~al.
\newblock Helpsteer: Multi-attribute helpfulness dataset for steerlm.
\newblock \emph{arXiv preprint arXiv:2311.09528}, 2023.

\bibitem[Wei et~al.(2023)Wei, Haghtalab, and Steinhardt]{wei2023jailbroken}
Wei, A., Haghtalab, N., and Steinhardt, J.
\newblock Jailbroken: How does {LLM} safety training fail?
\newblock In \emph{NeurIPS}, 2023.

\bibitem[Wong et~al.(2020)Wong, Rice, and Kolter]{wong2020fast}
Wong, E., Rice, L., and Kolter, J.~Z.
\newblock Fast is better than free: Revisiting adversarial training.
\newblock \emph{arXiv preprint arXiv:2001.03994}, 2020.

\bibitem[Xiao et~al.(2018)Xiao, Li, Zhu, He, Liu, and Song]{xiao2018generating}
Xiao, C., Li, B., Zhu, J.-Y., He, W., Liu, M., and Song, D.
\newblock Generating adversarial examples with adversarial networks.
\newblock In \emph{IJCAI}, 2018.

\bibitem[Xiao et~al.(2022)Xiao, Yang, Fan, Wang, and Luo]{xiao2022understanding}
Xiao, J., Yang, L., Fan, Y., Wang, J., and Luo, Z.-Q.
\newblock Understanding adversarial robustness against on-manifold adversarial examples.
\newblock \emph{arXiv preprint arXiv:2210.00430}, 2022.

\bibitem[Ye et~al.(2020)Ye, Gong, and Liu]{ye-etal-2020-safer}
Ye, M., Gong, C., and Liu, Q.
\newblock {SAFER}: A structure-free approach for certified robustness to adversarial word substitutions.
\newblock In \emph{ACL}, 2020.

\bibitem[Zang et~al.(2019)Zang, Qi, Yang, Liu, Zhang, Liu, and Sun]{zang2019word}
Zang, Y., Qi, F., Yang, C., Liu, Z., Zhang, M., Liu, Q., and Sun, M.
\newblock Word-level textual adversarial attacking as combinatorial optimization.
\newblock \emph{arXiv preprint arXiv:1910.12196}, 2019.

\bibitem[Zhang et~al.(2019{\natexlab{a}})Zhang, Zhang, Lu, Zhu, and Dong]{zhang2019you}
Zhang, D., Zhang, T., Lu, Y., Zhu, Z., and Dong, B.
\newblock You only propagate once: Accelerating adversarial training via maximal principle.
\newblock In \emph{NeurIPS}, 2019{\natexlab{a}}.

\bibitem[Zhang et~al.(2019{\natexlab{b}})Zhang, Yu, Jiao, Xing, El~Ghaoui, and Jordan]{zhang2019theoretically}
Zhang, H., Yu, Y., Jiao, J., Xing, E., El~Ghaoui, L., and Jordan, M.
\newblock Theoretically principled trade-off between robustness and accuracy.
\newblock In \emph{ICML}, 2019{\natexlab{b}}.

\bibitem[Zhang et~al.(2015)Zhang, Zhao, and LeCun]{zhang2015character}
Zhang, X., Zhao, J., and LeCun, Y.
\newblock Character-level convolutional networks for text classification.
\newblock In \emph{NeurIPS}, volume~28, 2015.

\bibitem[Zheng et~al.(2023)Zheng, Chiang, Sheng, Zhuang, Wu, Zhuang, Lin, Li, Li, Xing, et~al.]{zheng2023judging}
Zheng, L., Chiang, W.-L., Sheng, Y., Zhuang, S., Wu, Z., Zhuang, Y., Lin, Z., Li, Z., Li, D., Xing, E., et~al.
\newblock Judging llm-as-a-judge with mt-bench and chatbot arena.
\newblock \emph{Advances in Neural Information Processing Systems}, 36:\penalty0 46595--46623, 2023.

\bibitem[Zheng et~al.(2024)Zheng, Chiang, Sheng, Zhuang, Wu, Zhuang, Lin, Li, Li, Xing, et~al.]{zheng2024judging}
Zheng, L., Chiang, W.-L., Sheng, Y., Zhuang, S., Wu, Z., Zhuang, Y., Lin, Z., Li, Z., Li, D., Xing, E., et~al.
\newblock Judging llm-as-a-judge with mt-bench and chatbot arena.
\newblock In \emph{NeurIPS}, volume~36, 2024.

\bibitem[Zhong et~al.(2023)Zhong, Huang, Wettig, and Chen]{zhong2023poisoning}
Zhong, Z., Huang, Z., Wettig, A., and Chen, D.
\newblock Poisoning retrieval corpora by injecting adversarial passages.
\newblock \emph{arXiv preprint arXiv:2310.19156}, 2023.

\bibitem[Zhu et~al.(2020)Zhu, Cheng, Gan, Sun, Goldstein, and Liu]{Zhu2020FreeLB}
Zhu, C., Cheng, Y., Gan, Z., Sun, S., Goldstein, T., and Liu, J.
\newblock Freelb: Enhanced adversarial training for natural language understanding.
\newblock In \emph{ICLR}, 2020.

\bibitem[Zou et~al.(2023)Zou, Wang, Kolter, and Fredrikson]{zou2023universal}
Zou, A., Wang, Z., Kolter, J.~Z., and Fredrikson, M.
\newblock Universal and transferable adversarial attacks on aligned language models.
\newblock \emph{arXiv preprint arXiv:2307.15043}, 2023.

\end{thebibliography}
